\documentclass[conf]{new-aiaa}
\usepackage[utf8]{inputenc}

\usepackage{graphicx}
\usepackage{amsmath}
\usepackage{subcaption}
\usepackage{wrapfig}
\usepackage{bbm}
\usepackage{soul}

\usepackage{algorithm}
\usepackage{algorithmic}
\usepackage{multirow}

\usepackage{hyperref}

\newcommand{\vect}[1]
{\mathbf{#1}}

\usepackage[version=4]{mhchem}
\usepackage{siunitx}
\usepackage{longtable,tabularx}
\usepackage{soul}
\usepackage{todonotes}
\usepackage{svg}

\newcommand\copyrighttext{%
  \footnotesize Copyright~\textcopyright~2024 by the American Institute of Aeronautics and Astronautics, Inc. All rights reserved.
}
\newcommand\copyrightnotice{%
\begin{tikzpicture}[remember picture,overlay]
\node[anchor=south,yshift=10pt,align=center] at (current page.south) {\fbox{\copyrighttext}};
\end{tikzpicture}%
}

\title{Bayesian Data Augmentation and Training for Perception DNN in Autonomous Aerial Vehicles}

\author{Ashik E Rasul\footnote{
Equal contribution}\textsuperscript{\dag}}
\author{Humaira Tasnim\textsuperscript{*}\footnote{
Graduate Student, Mechanical Engineering Department, 1 William L Jones Dr, Cookeville, TN 38505.}}
\author{Hyung-Jin Yoon\footnote{
Assistant Professor, Mechanical Engineering Department, 1 William L Jones Dr, Cookeville, TN 38505.}}
\affil{Tennessee Technological University, Cookeville, TN 38505 USA}

\author{Ayoosh Bansal\footnote{Postdoctoral Research Associate, Computer Science, 201 North Goodwin Avenue, Urbana, IL 61801, USA.}}
\affil{University of Illinois at Urbana-Champaign, Urbana, Illinois, 61801}

\author{Duo Wang\footnote{Postdoctoral Research Associate, Mechanical Engineering Department, 1664 N. Virginia Street, Reno, NV 89557}
\footnote{Visiting Scholar at UIUC, Mechanical Science \& Engineering Department, 105 S Mathews Ave, Urbana, IL 61801}
}
\affil{University of Nevada, Reno, NV 89557 USA}

\author{Naira Hovakimyan\footnote{Professor, Mechanical Engineering Department }, Lui Sha\footnote{Donald B. Gillies Chair, Computer Science, 201 North Goodwin Avenue, Urbana, IL 61801, USA.}}
\affil{University of Illinois at Urbana-Champaign, Urbana, Illinois, 61801}

\author{Petros Voulgaris\footnote{Professor, Mechanical Engineering Department, 1664 N. Virginia Street, Reno, NV 89557.}}
\affil{University of Nevada, Reno, NV 89557 USA}

\begin{document}

\maketitle
\copyrightnotice

\begin{abstract}

Learning-based solutions have enabled incredible capabilities for autonomous systems.
Autonomous vehicles, both aerial and ground, rely on Deep Neural Networks (\emph{DNN}) for various integral tasks, including perception.
The efficacy of supervised learning solutions, such as the \emph{DNN} used for perception tasks, hinges on the quality of the training data.
Discrepancies between training data and operating conditions result in faults that can lead to catastrophic incidents.
However, collecting and labeling vast amounts of context-sensitive data, with broad coverage of possible variations in the operating environment, is prohibitively difficult.
To overcome this limitation, synthetic data generation techniques for \emph{DNN} training emerged, allowing for the easy exploration of diverse scenarios.
While significant synthetic data generation solutions exist for ground vehicles,
for aerial vehicles such support is still lacking.

This work presents a data augmentation framework for aerial vehicle's perception training, leveraging photorealistic simulation seamlessly integrated with high-fidelity vehicle dynamics, control, and planning algorithms.
Safe landing in urban environments is a crucial challenge in the development of autonomous air taxis, and therefore, landing maneuver is chosen as the focus of this work.
With repeated simulations of landing maneuvers in scenarios with varying vehicle states, weather conditions and time of day, we
assess the landing performance of the \emph{VTOL} (Vertical Take off and Landing) type UAV and gather valuable data. The landing performance is used as the objective function to optimize the \emph{DNN} through retraining. Given the high computational cost of \emph{DNN} retraining, we incorporated Bayesian Optimization in our framework that systematically explores the data augmentation parameter space to retrain the best-performing models.
The framework allowed us to identify high-performing data augmentation parameters that are consistently effective across different landing scenarios.
Utilizing the capabilities of this data augmentation framework, we obtained a robust perception model. The model consistently improved the perception-based landing success rate by at least 20\% under different lighting and weather conditions.

\end{abstract}

% =======================================================%
\section{Introduction}
% =======================================================%
Perception solutions for autonomous vehicles are mostly based on \emph{Deep Neural Network}s \emph{(DNN)}. Numerous studies have demonstrated their superior performance compared to traditional computer vision methods~\cite{grigorescu2020survey}. Specific frameworks like \emph{You Only Look Once (YOLO)}~\cite{redmon2016,redmon2017yolo9000,ultralytics2020yolov5,ultralytics2023yolov8} showcase the power of \emph{DNN}s in real-time object detection tasks. However, the success of \emph{DNN}
depends heavily on the quality and relevance of the training data. \emph{DNN}s trained exclusively on general datasets often show limited performance when applied to specific applications~\cite{Gupta_2024_WACV}.
This performance gap is rooted in the differences between the training data and the deployment environment~\cite{nvidia_synthetic}.
Optimizing performance in complex environments requires fine-tuning the model's hyperparameters by retraining it within the deployment environment.

Optimizing a \emph{DNN} model for deployment in complex real-world environments comes with two significant challenges.
First, such retraining requires large amounts of data collected in varying scenarios.
For aerial vehicle perception, this would involve the use of expensive aerial vehicles over long periods of time to cover a comprehensive range of relevant scenarios.
This limitation can be alleviated by using photorealistic simulators to freely explore
various scenarios and collect task-specific data~\cite{nvidia_synthetic, dallel2023digital, unity_autonomous_vehicle_training}.
Second, retraining \emph{DNN} models is computationally expensive, requiring substantial resources and time to refine them for real-world applications~\cite{thompson2020computational}.
This high computational demand often becomes a bottleneck in the optimization process, necessitating improvements in the efficiency of the retraining process itself.
Bayesian optimization has become a prominent method for this purpose, as it systematically searches the decision variable space, balancing exploration and exploitation to identify optimal decision variables without knowing the objective function explicitly~\cite{frazier2018tutorial}. This approach is particularly effective for deep neural network (\emph{DNN})-based perception systems, where robust performance depends on selecting hyperparameters that generalize well across diverse scenarios~\cite{victoria2021automatic}.
By leveraging feedback from the simulation outcomes, such as landing success rates, Bayesian optimization iteratively refines the model, ensuring it adapts effectively to varying vehicle and environmental states.

\begin{figure}[t]
    \centering
    \includegraphics[width=0.95\textwidth]{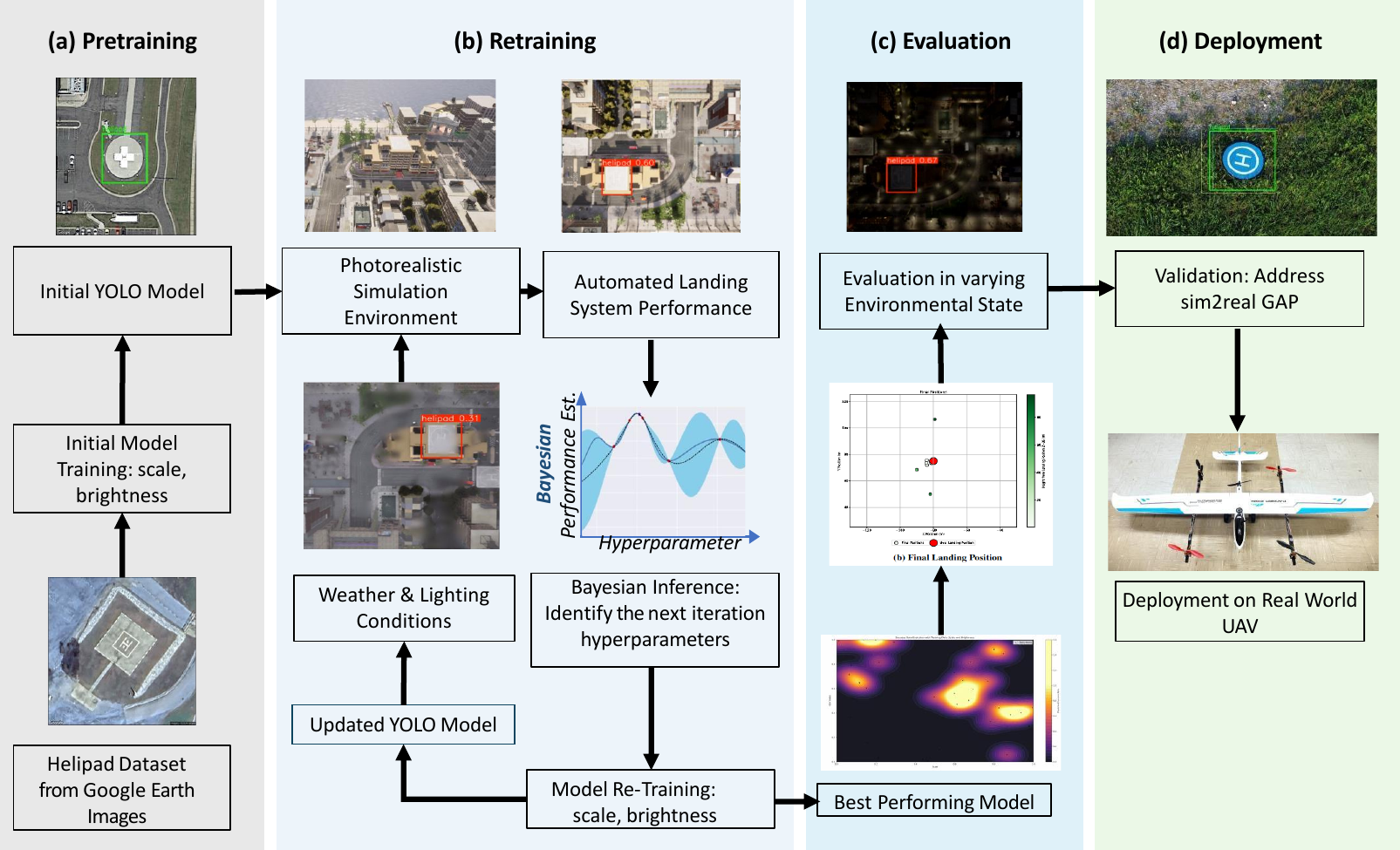}
    \caption{Data Exploration Framework
    }
    \label{fig:framework}
\end{figure}

In this work, we address both challenges mentioned above in the context of perception systems for autonomous aerial vehicles, such as air taxis. Specifically, we focus on optimizing a \emph{DNN}-based perception model for an automated landing system on a \emph{Vertical Take-Off and Landing (VTOL)} aerial vehicle.
The autonomous landing system employs a \emph{DNN} object detector to identify the landing spot within a camera image frame. We propose an automated decision-making framework designed to enhance the utility of data samples collected for retraining, thereby improving the performance of the landing pad detector.
This framework consists of two iterative phases. The first phase evaluates the overall performance of the vision-guided landing system and the uncertainties of the computer vision model given the operation scenarios in photorealistic simulation. In the second phase, the vehicle states and corresponding measure of cumulative landing success rate for the system are used in a Bayesian optimization loop to generate the next set of hyperparameter for retraining the model. Finally, the updated model is evaluated in the first phase. These phases close the loop in the data exploration process, as shown in Figure~\ref{fig:framework}. To the best of our knowledge, this is the first use of such a framework for vision-based landing maneuvers in \emph{VTOL}.
Our key contribution can be summarized as follows:

\begin{itemize}
    \item We develop a Bayesian retraining framework\footnote{\url{https://github.com/arasul42/LCASL-TTU-vtol-bayes-sim}} for the \emph{YOLO} perception model~\cite{redmon2017yolo9000} for vision-based autonomous landing within a photorealistic simulation environment based on the \emph{CARLA} simulator~\cite{Dosovitskiy17}, incorporating high-fidelity aerial vehicle dynamics from the \emph{GUAM} vehicle simulation~\cite{GUAM2024}.
\end{itemize}

\section{Literature Review}

Autonomous landing systems in \emph{VTOL} vehicles depend on reliable landing pad detection.
However, this detection is challenged by a wide range of variable environmental factors, such as weather conditions and lighting.
Real-world datasets, particularly those for aerial vehicles, are unable to capture the full combinatorial complexity of all the factors that can impact the performance of landing pad detection.
Synthetic data generation and data augmentation techniques have emerged as essential tools in the development of machine learning models to address this limitation of real-world datasets.
The simulation-based methods enable the creation of diverse and customizable datasets, leading to improved resilience of models trained using these datasets.
This section provides a review of related works on the components integrated into our proposed framework, including data augmentation, Bayesian optimization, and the use of \emph{\emph{DNN}s} in control applications.

\begin{itemize}
    \item \textbf{\textit{Synthetic Data and Data Augmentation}}: Synthetic data generation and data augmentation have become indispensable tools in enhancing the robustness of perception models for \emph{VTOL}s.
    Docca and Torculas~\cite{nvidia_synthetic} highlight that synthetic data generated through simulation environments enables the creation of diverse image scenarios without the need for extensive manual data collection.
    These datasets enable model training on a wide variety of visual conditions. This helps the model generalize effectively in dynamic real-world environments.
    \emph{Unity}’s simulated environments~\cite{unity_autonomous_vehicle_training} contribute further by replicating complex landing scenarios and lighting conditions that are essential for robust \emph{VTOL} perception. However, despite its flexibility,
    \emph{Unity}'s simulated environments are not suitable for replicating the full complexity of real-world \emph{VTOL} dynamics, as \emph{Unity} is primarily designed for \emph{VR} and game applications.
    Accurately modeling sensor noise, unpredictable environmental changes, and \emph{VTOL} hardware limitations are the key difficulties that are crucial during operations like autonomous landings. While \emph{Unity} excels in visual fidelity, the gap between simulated physics and real-world interactions necessitates further validation through real-world data or hybrid approaches combining simulations with digital twins~\cite{dallel2023digital}.

    \item \textbf{\textit{Weather Augmentation}}: Building on the strengths of synthetic data, weather augmentation adds a targeted approach to improving model robustness under adverse conditions. By generating realistic weather variations, such as rain, fog, and snow, during training, models become more robust and adaptable. Gupta et al.~\cite{Gupta_2024_WACV} highlight the importance of generating realistic weather variations, such as rain and fog, during training to enhance model robustness. By introducing weather-based transformations into datasets, models can generalize better to challenging real-world scenarios. Furthermore, techniques like adversarial augmentation add further complexity, which simulate edge cases that are difficult to replicate in physical experiments~\cite{xie2019ada}.

    \item \textbf{\textit{Digital Twins and Virtual Reality Environments}}: While synthetic data focuses on creating individual variations, digital twins and Virtual Reality (\emph{VR}) environments provide dynamic simulations of operational scenarios. Digital twins go beyond static synthetic data by simulating operational scenarios dynamically and integrating real-time environmental factors like wind and temperature~\cite{dallel2023digital}. Ensuring the fidelity of these twins, however, remains a challenge, especially for \emph{VTOL}s operating in rapidly changing conditions.
    While synthetic data generates isolated variations, digital twins provide a comprehensive simulation of the \emph{VTOL}'s operational environment, incorporating variables such as lighting, weather, and terrain~\cite{dallel2023digital}.
    By leveraging these technologies, \emph{VTOL} perception systems gain robustness, improving model reliability in dynamic environments. However, the fidelity of \emph{VR} simulations can sometimes differ from real-world conditions, underscoring the need for iterative improvement~\cite{mishra2024closing}.

    \item \textbf{\textit{Efficient Data Selection Techniques}}: As the complexity of \emph{VTOL} perception tasks grows, data selection becomes crucial in optimizing model training. Selecting the most informative data points ensures that the model focuses on samples that are difficult to learn, reducing both training time and computational load. Reducing the computational load of large-scale \emph{VTOL} perception systems is essential. Techniques like \emph{coreset} selection~\cite{paul2021deep} and prioritized training~\cite{Gupta_2024_WACV} construct efficient datasets that balance size and representativeness. Mindermann et al.~~\cite{mindermann2022prioritized} introduce a technique for prioritizing \emph{learnable} and \emph{informative} points, emphasizing that concentrating on challenging data can, in fact, enhance model accuracy. Similarly, Paul et al.~\cite{paul2021deep} highlight \emph{data pruning} techniques, which refine the training set by eliminating redundant examples, helping the model converge faster. Advanced data selection methods, such as those discussed by Guo et al.~\cite{guo2022deepcore} and Sorscher et al.~\cite{sorscher2022beyond}, use \emph{coreset} selection to construct compact, representative datasets. These methods are especially valuable in \emph{VTOL} perception, where reducing dataset size without compromising performance is crucial for model improvement. By prioritizing efficient data selection, \emph{VTOL} models can maintain high performance while minimizing computational resource requirements.

    \item \textbf{\textit{Bayesian Inference and Optimization}}: In complex \emph{VTOL} perception systems, Bayesian inference offers a structured approach to quantify and incorporate uncertainty into model predictions, ultimately enhancing model robustness and adaptability. This approach is particularly beneficial in dynamic scenarios like \emph{VTOL} landing, where conditions continuously evolve. Deng et al.~\cite{deng2024towards} propose Bayesian data selection methods that leverage uncertainty estimates to refine training datasets, focusing on data points that improve model learning efficiency. Additionally, Shahriari et al.~\cite{shahriari2015taking} provide a comprehensive review of Bayesian optimization techniques, underscoring its potential to iteratively adjust model parameters for better performance.
    For the \emph{VTOL} landing tasks addressed in this work, we utilized Bayesian optimization~\cite{snoek2012practical} to fine-tune critical perception parameters, such as scale and brightness, achieving substantial improvements in detection accuracy across a wide range of environmental conditions.

    \item \textbf{\textit{Simulation Environments and Object Detection}}:
    Advanced simulation environments like \emph{CARLA} and object detection models like \emph{YOLO} provide the infrastructure for rigorous testing of \emph{VTOL} systems. \emph{CARLA} offers detailed urban and rural scenarios with adjustable parameters such as lighting, weather, and environmental objects, enabling \emph{VTOL}s to be tested in diverse landing contexts~\cite{Dosovitskiy17}. \emph{YOLO}, known for its speed and accuracy in object detection, is particularly suited for real-time applications. Redmon and Farhadi’s \emph{YOLO9000} model~\cite{redmon2017yolo9000} and subsequent iterations have become widely used for \emph{VTOL} applications, offering efficient and reliable detection capabilities in \emph{CARLA}’s complex simulated environments. By integrating \emph{YOLO} models in \emph{CARLA}, \emph{VTOL} systems can gain real-world insights and refine detection accuracy under various conditions. This combination of \emph{CARLA}’s simulation environment and \emph{YOLO}’s detection capabilities represents a powerful toolkit for advancing safe \emph{VTOL} landing technology.

    \item \textbf{\textit{Uncertainty Quantification and Adaptive Control}}: Uncertainty quantification (\emph{UQ}) enhances \emph{VTOL} decision making by identifying high-confidence predictions, ensuring safety in dynamic conditions. \emph{UQ} is critical in \emph{VTOL} perception and control since it allows the system to estimate the reliability of its predictions in dynamic, often unpredictable environments. \emph{UQ} helps identify high-confidence detections and ensures the \emph{VTOL} only proceeds with safe, verified actions. Fernández Castaño et al. ~\cite{fernandez2024uncertainty} demonstrate a \emph{UQ}-based switching control approach that enables \emph{VTOL}s to adapt their behavior based on environmental conditions, enhancing tracking accuracy. Adaptive control, together with \emph{UQ}, offers a proactive response to uncertainty by adjusting control parameters in real time. For instance, Cao and Hovakimyan’s $\mathcal{L}_1$ adaptive control architecture~\cite{cao2008design} emphasizes rapid adaptation, ensuring stability and performance despite unpredicted changes. By combining \emph{UQ} with adaptive control, \emph{VTOL}s become better equipped for precise, reliable landing operations, even under uncertain conditions.

    \item \textbf{\textit{Neural Network-Based Control Methods}}: Recent advancements in neural network-based control methods have shown promise in optimizing complex dynamical systems without the need for extensive retraining. Hose et al.\cite{hose2024finetuning} introduce a method for fine-tuning Approximate Model-Predictive Control (\emph{AMPC}) using Bayesian Optimization (\emph{BO}) to adapt neural network controllers efficiently. The proposed technique addresses a key challenge in \emph{AMPC}: the need for retraining when system parameters change. By leveraging \emph{BO}, the method allows for data-driven adaptation of \emph{AMPC}, reducing manual tuning efforts and computational costs. The authors validated their approach on challenging control tasks, such as the cart-pole system and a balancing unicycle robot, demonstrating improved stability and reduced oscillations with minimal hardware experimentation. In a related approach, Ostafew et al.\cite{6907444} applied a learning-based nonlinear Model-Predictive Control (\emph{MPC}) to enhance vision-based path tracking for mobile robots in challenging outdoor environments. The study highlights the importance of integrating vision-based perception with advanced control strategies to handle unpredictable conditions. This technique utilized data-driven learning to adjust the controller in real-time based on visual feedback, significantly improving path-tracking performance. The insights from Ostafew et al.'s work are directly applicable to \emph{VTOL} landing perception, where the integration of vision-based inputs and adaptive control methods is crucial for accurately detecting and landing on helipads in dynamic and uncertain environments.

    \end{itemize}

The literature highlights the importance of a comprehensive approach that combines synthetic data generation, weather augmentation, advanced data selection techniques, simulation environments, and uncertainty-aware modeling to enhance \emph{VTOL} perception systems. In particular, weather augmentation bridges the gap between idealized training conditions and real-world variability, ensuring \emph{VTOL}s can operate reliably in diverse and dynamic environments. By integrating these advancements, future \emph{VTOL} systems will achieve increased robustness and operational safety, enabling dependable autonomous landings even in challenging conditions.

%%%%%%%%%%%%%%%%%%%%%%%%%%%%%%%%%%%%%%%%%%%%%%%%%%%%%%
\section{Proposed Framework}
%%%%%%%%%%%%%%%%%%%%%%%%%%%%%%%%%%%%%%%%%%%%%%%%%%%%%%
Developing perception systems for autonomous \emph{VTOL}s demands a structured approach to address challenges such as data variability, hyperparameter optimization, and performance evaluation across diverse conditions. Our proposed framework addresses these challenges by integrating data augmentation, high-fidelity simulation environments, uncertainty quantification, and Bayesian optimization to enhance the performance of vision-based \emph{VTOL} landing systems.
This approach aims to iteratively adapt the model to new environments and enhance the landing success rate through systematic evaluation and retraining.
The proposed framework, illustrated in Figure~\ref{fig:framework}, comprises five main components: \textbf{Dataset Processing and Initial Model Training}, which involves preparing the dataset and training an initial \emph{YOLO} model with reproducible results for helipad detection; \textbf{Uncertainty Estimation}, where prediction uncertainties are quantified to identify areas for model improvement and guide hyperparameter tuning; \textbf{Simulation Testing}, which evaluates the trained model in a high-fidelity, photorealistic simulation environment that incorporates vehicle dynamics; \textbf{Retraining Loop with Bayesian Optimization}, which iteratively refines the model by optimizing hyperparameters based on simulation feedback; and finally, \textbf{Validation Under Adverse Conditions}, where the model's robustness is tested under diverse weather and lighting scenarios to ensure reliable performance.
The components are described below:

% ================================================================%
\subsection{Dataset Processing and Initial Model Training}
% ================================================================%

The importance of working with a high-quality dataset in training a model cannot be overstated, as the dataset fundamentally determines the model's learning and performance~\cite{nvidia_synthetic,deng2024towards}.
Therefore,
the first component of our framework focuses on data preprocessing and augmentation to train reproducible image detection \emph{YOLO} models, specifically \emph{YOLOv8}~\cite{terven2023yolo}. As dataset, we use \emph{Google Earth} images of helipads~\cite{bitoun2020helipadcat}. \emph{YOLO}'s fast inference time and efficiency make it well-suited for detecting objects in challenging real-time environments on mobile platforms~\cite{redmon2017yolo9000}. \emph{YOLO} models, such as \emph{YOLOv5} and \emph{YOLOv8}, employ random data augmentations during training to enhance model robustness and generalization. These augmentations include techniques like random resizing, flipping, cropping, color and brightness adjustments. By default, the randomness inherent in these augmentations leads to non-reproducible results across different training runs~\cite{ultralytics2024}.

Reproducibility in empirical AI research is the ability of an independent research team to produce the same results using the same AI method based on the documentation made by the original research team~\cite{tatman2018reproducibility}. Highly reproducible training ensures the same results are obtained over repeated iterations with the same environment, code and data~\cite{tatman2018practical}. To achieve reproducibility, we turn off the random augmentation of \emph{YOLO} training, which gives us deterministic control over the image augmentation during the training cycle.
This approach also allows us to isolate the effects of specific hyperparameters and observe the model's behavior on unseen data.
However, this method of hyperparameter tuning has the inherent risk of overfitting the model that performs well only on that specific scenario, reducing its ability to generalize to unseen data.
Our framework reduces the risk of overfitting by validating model performance under diverse lighting and weather conditions.

The training dataset~\cite{bitoun2020helipadcat} contains coordinates of helipads of different types, including their \emph{Google Earth} URL, image size details, and ground truth bounding box annotations.
The final dataset consists of approximately 4,000 images in JPEG format, each with a resolution of 640×640 pixels.
The dataset labels are provided as the minimum and maximum coordinates of the bounding box (ground truth), which are not directly compatible with the open-source \emph{YOLO} implementation used in this work~\cite{YOLOv8}.
\emph{YOLO} requires annotations in a specific text file format, where bounding box coordinates are normalized relative to the image dimensions~\cite{redmon2017yolo9000}.
We transformed the label annotations to a \emph{YOLO}-compatible format to prepare the dataset. Each annotation is stored in a plain text file corresponding to the image file, with rows representing individual objects and specifying their respective classes.
Each annotation follows the format: \texttt{<class\_id> <x\_center> <y\_center> <width> <height>}
where each value is normalized as per the \emph{YOLO} format. Here, \texttt{<class\_id>} represents the object class, which is helipad in our case, \texttt{<x\_center>} and \texttt{<y\_center>} are the normalized coordinates of the bounding box centre relative to the image width and height and \texttt{<width>} and \texttt{<height>} are the normalized dimensions of the bounding box.
After preprocessing, each image was paired with its corresponding annotation file in a directory, preparing the dataset for \emph{YOLO} object detection training.

\begin{figure}[h!]
    \centering
    \begin{subfigure}{0.45\textwidth}
        \centering
        \includegraphics[width=0.95\textwidth]{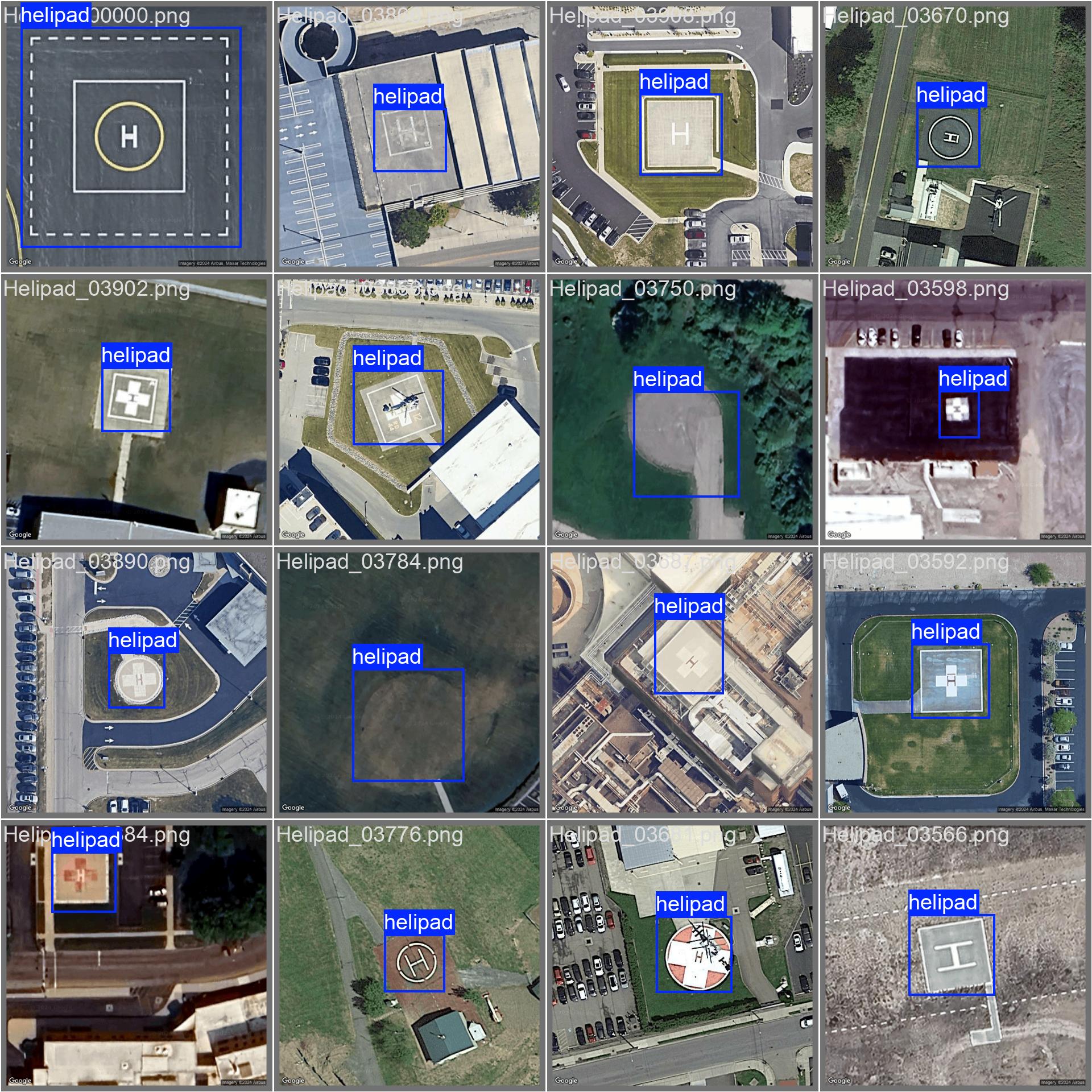}
        \caption{Helipad Dataset}
        \label{fig:helipad}
    \end{subfigure}
    \hspace{0.02\textwidth}
    \begin{subfigure}{0.50\textwidth}
        \centering
        \includegraphics[width=\textwidth]{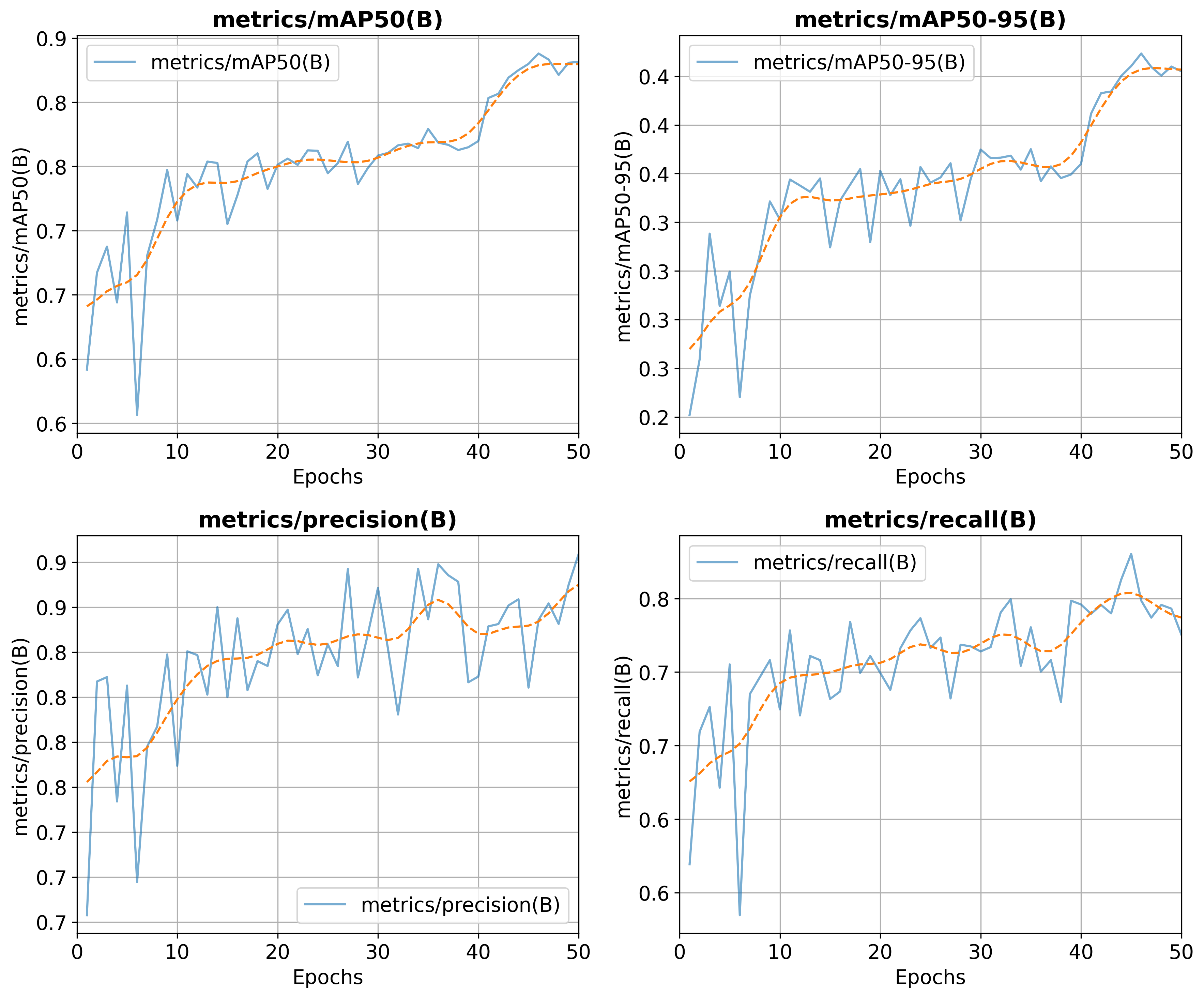}
        \caption{\emph{YOLO} Training Result
        }
        \label{result}
    \end{subfigure}
    \caption{Initial Model Training}
    \label{fig:training_resutl}
\end{figure}

From the dataset, 80\% of the images were randomly selected for training, while the remaining 20\% were reserved for validation. Figure~\ref{fig:helipad} shows sample training images annotated with the \emph{helipad} class bounding boxes, demonstrating the diversity and distribution of the dataset. The \emph{YOLO} model outputs bounding box coordinates with confidence scores, indicating the likelihood of correct helipad detection. The model’s performance was evaluated using standard object detection metrics, including mean Average Precision (mAP) at IoU (Intersection over Union) thresholds of 50\% (mAP50) and a range from 50\% to 95\% (mAP50-95), as shown in Figure~\ref{result}. These metrics provide a comprehensive assessment of the model's detection accuracy. IoU measures the overlap between the predicted bounding box and the ground truth bounding box, serving as a criterion for correct detection~\cite{lin2014microsoft}. Specifically, IoU is defined as the ratio of the intersection area to the union area of the two bounding boxes, with higher values indicating better alignment between the prediction and the ground truth~\cite{lin2014microsoft}. Precision at a single IoU threshold (50\%) is measured by mAP50, indicating how well the model detects objects at a basic level of overlap and mAP50-95 averages precision over multiple IoU thresholds~\cite{lin2014microsoft}. This metric mAP50-95 provides a more rigorous evaluation of the model's performance, especially for scenarios requiring fine-grained object detection~\cite{lin2014microsoft}. The curves in the top two charts illustrate that as training progresses over epochs, both mAP50 and mAP50-95 steadily increase, indicating improvements in the model's ability to detect helipads accurately. The precision and recall metrics, shown in the bottom two charts, provide further insight into the model’s performance. Precision evaluates the proportion of correctly detected helipads out of all detections made, reflecting the model’s ability to avoid false positives~\cite{lin2014microsoft}. Recall measures the proportion of correctly detected helipads out of all actual helipads, indicating how well the model avoids false negatives~\cite{lin2014microsoft}. These trends highlight the model's increasing generalization capability and robustness when applied to the validation set.

% ================================================================%
\subsection{Uncertainty Estimation}
\begin{figure}[hbt!]
     \centering
     \begin{subfigure}[b]{0.65\columnwidth}
         \centering
         \includegraphics[width=\columnwidth]{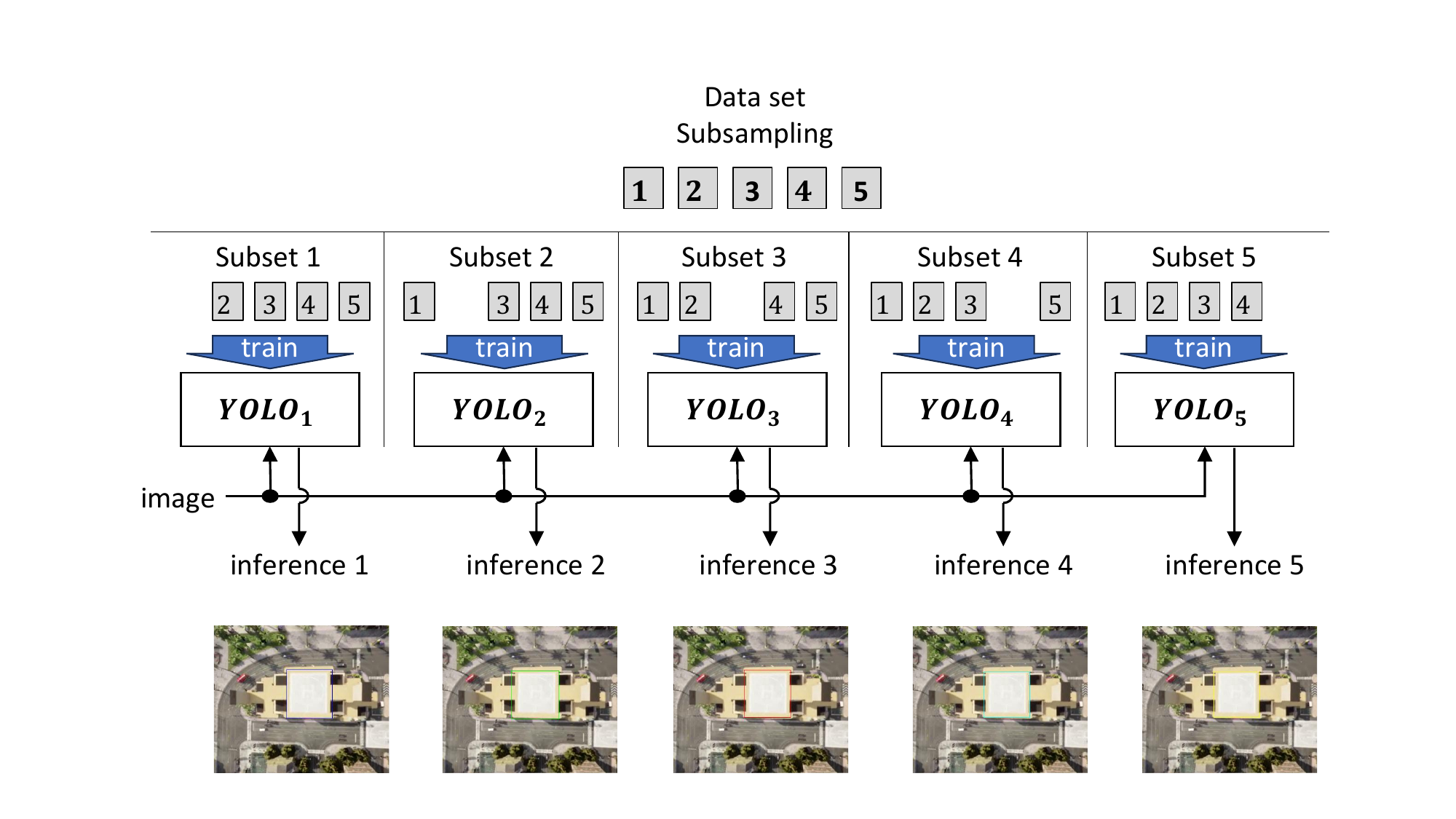}
         \caption{Dataset Bootstrapping to Train Models}
         \label{subsampling}
     \end{subfigure}
     \hfill
     \begin{subfigure}[b]{0.30\columnwidth}
         \centering
         \includegraphics[width=\columnwidth]{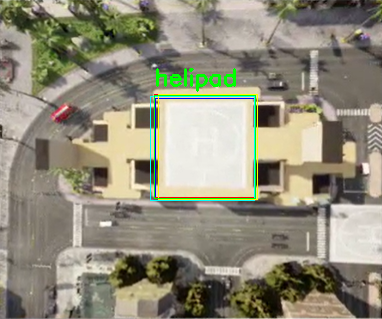}
         \caption{Disagreement among models}
         \label{day}
     \end{subfigure}
     \hfill
     \caption{Uncertainty Estimation in The System}
     \label{Uncertainty}
\end{figure}

% ================================================================%
The initial model training process was repeated five times using dataset \emph{Subsampling}, similar to the approach used in random forests~\cite{breiman2001random}.
\emph{Subsampling} is a method where subsets of data are drawn randomly, without replacement, from the original dataset. This method allows the creation of multiple training datasets smaller than the original, effectively simulating the variability of data distributions. It is beneficial while working with small datasets, as it provides a way to estimate the model’s robustness and reduces over-fitting by introducing diversity into training subsets. First, we separate the validation dataset (20\% of the total dataset). Second, we randomly distribute the remaining 80\% of the dataset to create five subsets of training data. For each \emph{YOLO} model training, 4 out of 5 subsets are combined and used as the training dataset, as shown in Figure~\ref{subsampling}. However, the validation set is kept the same to ensure consistency in evaluation.
For each training step, the \emph{YOLO} model was trained to detect helipads, producing five bounding boxes corresponding to the class \emph{helipad}. The standard deviation of the five predicted bounding box coordinates provides a measure of uncertainty, as illustrated in Figure~\ref{day}. This uncertainty represents the epistemic uncertainty~\cite{huellermeier2019aleatoric} of the system, which is observed to be significant under lighting and weather conditions different from those in the training dataset.

% ================================================================%
\subsection{Simulation Testing}
% ================================================================%
The trained \emph{YOLO} model is integrated into the high-fidelity photorealistic simulation environment \emph{CARLA}, to evaluate its performance in controlled yet dynamic conditions. \emph{CARLA}, built using the \emph{Unreal Engine}, provides a photorealistic simulation of urban and rural environments, making it an ideal platform for testing autonomous systems~\cite{Dosovitskiy17,Unreal}. \emph{CARLA} also lets us define many environmental conditions, which in turn lets us generate realistic-looking challenging conditions~\cite{Dosovitskiy17}. This flexibility facilitates a safe and repeatable testing environment without the risks associated with real-world \emph{VTOL} operations, especially during the early stages of development~\cite{Dosovitskiy17}. The simulation specifically assessed the model's ability to detect helipads.
The photorealistic rendering closely mimicked real-world challenges, providing a comprehensive testbed to evaluate the \emph{YOLO} model's robustness. The next step involves simulating vehicle dynamics.

\emph{Generic Urban Air Mobility (GUAM)}, developed by \emph{NASA}, is a simulator designed for common rigid body 6-DOF (Degree of Freedom) frames of reference. It includes a wide range of aerospace signals and quantities of interest. The simulator's modular architecture readily supports the integration and replacement of aircraft models, sensors, actuator models, and control algorithms~\cite{GUAM2024}.
\emph{GUAM} simulation is migrated from \emph{MATLAB Simulink} to a \emph{Python} environment using \emph{Google}’s \emph{JAX} framework in~\cite{jaxguam2023}. \emph{JAX} leverages GPU parallel computing and provides automatic differentiation, similar to \emph{PyTorch}~\cite{NEURIPS2019_9015}, enabling faster simulation speeds of up to 200 Hz.
To integrate \emph{CARLA} with \emph{JAX-GUAM} and the \emph{YOLO} detection model, we utilized \emph{ROS}~\cite{ros} as middleware. Once the systems are connected, the next step involves implementing a control algorithm. In \emph{CARLA}, we mounted a downward-facing camera at the bottom of the aerial vehicle. The images captured by this camera serve as input for our landing system.

\begin{figure}[h!]
    \centering

    \begin{subfigure}{0.49\textwidth}
        \centering
        \includegraphics[width=\textwidth]{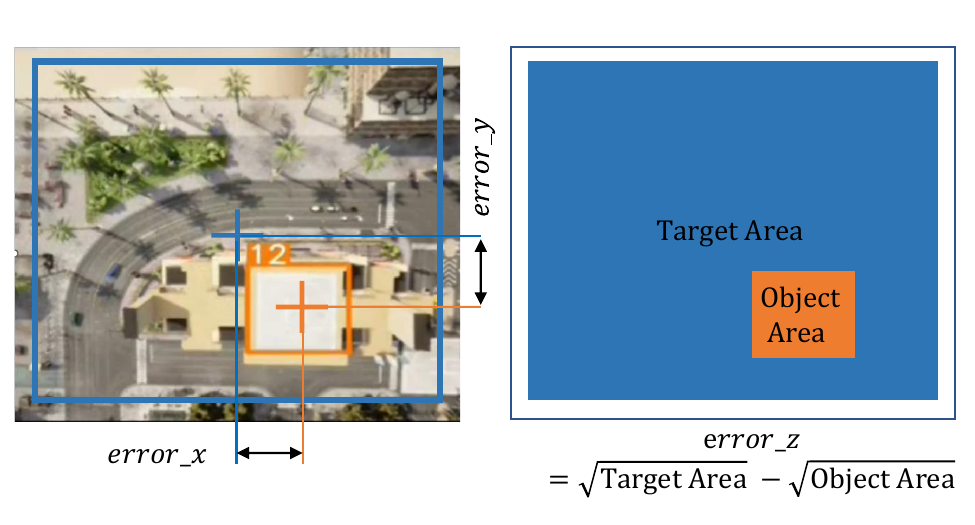}
        \caption{Error generation for the controller}
        \label{fig:error}
    \end{subfigure}
    \begin{subfigure}{0.49\textwidth}
        \centering
        \includegraphics[width=\textwidth]{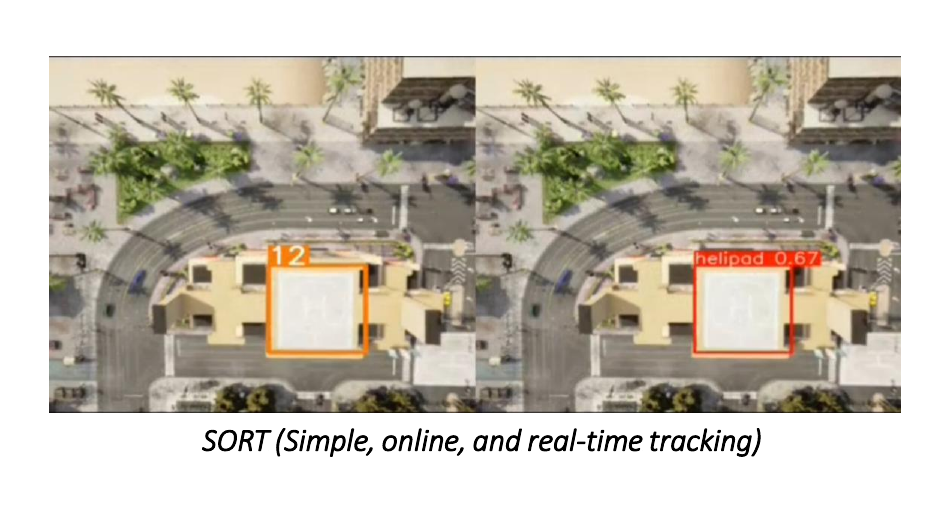}
        \caption{SORT implementation}
        \label{fig:sort}
    \end{subfigure}
    % Second image

    \hspace{0.02\textwidth}
    \caption{Object Detection, Tracking and Landing Controller }
    \label{fig:combined}
\end{figure}

The bottom camera of the vehicle captures image frames that are streamed into the \emph{YOLO} model to generate bounding boxes for the detected helipad.
The position and size of the detected bounding box are compared to the camera's center bounding box (the target area), generating an error vector, as shown in Figure~\ref{fig:error}. The camera center bounding box is represented by the blue box in the figure. To handle multi-object detection, we implemented the \emph{SORT (Simple Online and Real-time Tracking)} method~\cite{bewley2016sort}, as illustrated in Figure~\ref{fig:sort}. \emph{SORT} is a multiple-object tracking algorithm specifically designed for video sequences. It ensures consistent tracking of the correct helipad bounding box in dynamic environments, even when the target is partially occluded or temporarily lost. With accurate object detection and \emph{SORT}, the \emph{VTOL} can continuously produce the necessary error signal for the control algorithm.

We define the landing control error as follows: $\vect{e}=\vect{b}_c - \vect{b}_o$
where $\vect{b}_c$ is the camera center bounding box and $\vect{b}_o$ is the detected bounding box. Both the $\vect{x}$ and $\vect{y}$ components of the error can be found from the bounding box center deviation in the respective axis. For the $\vect{z}$ component of the error, we take the size difference of the object compared to the target area (Camera Center). This error is high when \emph{VTOL} is at a high altitude and low at the landing position. This method works if the landing pad size is less than the camera center bounding box at the landing position, which is 50~$\text{ft}^2$ in our case.
This error is processed by a PID controller~\cite{aastrom2021feedback} to compute the corrective velocity vector, $\vect{v}_{\text{corr}}=[v_x, v_y, v_z]^\top$:
\begin{equation*}
   v_i = K_p e_i + K_i \int e_i \, dt + K_d \frac{de_i}{dt}, \quad i \in \{x, y, z\},
\end{equation*}
where \(v_i\) is the corrective velocity along axis \(i\), \(e_i\) is the error along axis \(i\), and \(K_p\), \(K_i\), \(K_d\) are the proportional, integral, and derivative gains for the respective axis~\cite{csumnu2017simulation}.
The calculated velocity commands are constrained to prevent unmanageable overshoot and then fed into \emph{JAX-GUAM}, which updates the vehicle's state, guiding it closer to the ideal landing position based on the corrective velocity vector $v_i$.

The landing performance evaluation consists of ten trials with random initial positions. A landing is considered successful if the vehicle is within 4 meters of the ideal center on the horizontal plane and within 1 meter on the $z$-axis. The success rate is calculated based on the number of successful landings out of 10 attempts. For example, if six trials are successful, the success rate is 60\%.

% ================================================================%
\subsection{Bayesian Optimization for Iterative Retraining of Perception \emph{DNN}}
% ================================================================%

The Bayesian optimization in this work aims to improve the landing success rate by iteratively retraining the object detection \emph{DNN} (\emph{YOLO}). In each iteration, the retraining hyperparameters are treated as decision variables in the optimization process. Unlike typical optimization problems that assume a known objective function or gradient, Bayesian optimization handles an unknown objective function, mapping hyperparameters to the landing controller’s performance.
In the Bayesian optimization, we estimate the unknown objective function, which evaluates the landing success rate, as a posterior distribution based on data samples of the hyperparameters and their corresponding objective function values (i.e., landing success rates). The hyperparameters considered include the scale factor and brightness value used in data augmentation during training data processing for each retraining iteration. As described in Algorithm~\ref{alg:bayesian_optimization}, we use the acquisition function to identify the most promising data point to explore. The acquisition function balances the trade-offs between exploration and exploitation. This approach is particularly useful for optimizing black-box objective functions, making it highly effective for identifying the optimal combination of hyperparameters and augmentation strategies for an unknown objective function $f(\mathbf{x})$~\cite{shahriari2015taking}. In our case, $\mathbf{x}$  represents the hyperparameters and the augmentation parameters, such as scale and brightness.

The optimization aims to find the performance maximizer as follows:
\begin{equation*}
    \mathbf{x}^* = \arg\max_{\mathbf{x} \in \mathcal{X}} f(\mathbf{x}),
\end{equation*}
where $\mathcal{X}$ is the space of possible hyperparameters and augmentation combinations and $f(\mathbf{x})$ is the landing success rate obtained from the photorealistic simulation environment.

We use the Gaussian Process (GP) to determine the posterior distribution given the data points obtained in data exploration. To use GP to model the unknown function $f(\mathbf{x})$, we assume $f(\mathbf{x}) \sim \mathcal{GP}(m(\mathbf{x}), k(\mathbf{x}, \mathbf{x}'))$. Here $m(\mathbf{x})$ is the mean function which is defined as $m(x) = \mathbb{E}[f(\mathbf{x})]$. $\mathbb{E}[f(\mathbf{x})]$ is the mean value that the function \( f(\mathbf{x}) \) is assumed to take, given the prior distribution or as inferred from the data. If no prior information is available, the mean function $m(\mathbf{x})$ is often set to zero for simplicity. And, $k(\mathbf{\mathbf{x}, \mathbf{x}'})$ is the covariance function (kernel), which defines the relationship between different combinations of hyperparameters~\cite{shahriari2015taking}.

After we set the upper and lower bounds for scaling and brightness hyperparameters. The next sample point is chosen based on an acquisition function $a(\mathbf{x})$, which balances exploration (sampling uncertain regions) and exploitation (sampling regions likely to yield high rewards):
\begin{equation*}
  \mathbf{x}_{n+1} = \arg\max_{\mathbf{x} \in \mathcal{X}} a(\mathbf{x}; \mathcal{D}_n)
\end{equation*}\
where $\mathcal{D}_n = \{(\mathbf{x}_i, f(\mathbf{x}_i))\}_{i=1}^n$ is the set of previous observations~\cite{snoek2012practical}. One of the most widely used and simplest acquisition functions is the mean function plus the upper confidence bound (UCB). Where,
\begin{equation*}
  UCB(\mathbf{x}) = \mu(\mathbf{x}) + \kappa \cdot \sigma(\mathbf{x})
\end{equation*}
where \( \mu(\mathbf{x}) \) is the posterior predictive mean, \( \sigma(\mathbf{x}) \) is the posterior predictive standard deviation, and \(\kappa\) controls the exploration/exploitation trade-off. In our case, we used \(\kappa = 2.567\), which encourages exploration by optimistically selecting new data points with high uncertainty, while also prioritizing areas with a high probability of finding the optimal solution. However, the value of \(\kappa\) can be adjusted to make the algorithm more exploratory or exploitative, depending on the desired balance.

After evaluating $f(\mathbf{x}_{n+1})$, the GP is updated to incorporate the new observation~\cite{rasmussen2010gaussian} into the data set:
\begin{equation*}
    \mathcal{D}_{n+1} = \mathcal{D}_n \cup \{(\mathbf{x}_{n+1}, f(\mathbf{x}_{n+1}))\}.
\end{equation*}
Note that the GP is a non-parametric model; therefore, updating the dataset corresponds to updating the GP itself.
The process is repeated until a stopping condition is met as:
\begin{equation*}
    \text{Stop if: } \max a(\mathbf{x}; \mathcal{D}_n) < \epsilon,
\end{equation*}
where $\epsilon$ is a predefined threshold~\cite{snoek2012practical}.

The Gaussian Process (GP) and acquisition function in Algorithm~\ref{alg:bayesian_optimization} were implemented using the \emph{Python} package for Bayesian Optimization~\cite{BayesinaOptPython}.

\begin{algorithm}
\caption{Bayesian Optimization for Scale and Brightness}
\label{alg:bayesian_optimization}
\begin{algorithmic}[1]
\STATE{\textbf{Given:} Parameter space $\mathcal{X} = \{(S, B) : S, B \in [0, 1]^2\}$}
\STATE{\textbf{Given:} Success rate function $f(S, B) = \text{EvaluateModel}(\text{TrainModel}(S, B))$}
\STATE{\textbf{Given:} Surrogate function $\hat{f}(S, B)$ modeled as $\mathcal{GP}(m(x), k(x, x'))$}
\STATE{\textbf{Given:} Acquisition function $\alpha(S, B)$ (e.g., Upper Confidence Bound)}
\STATE{\textbf{Given:} Initial samples $\mathcal{X}_0 \leftarrow \text{RandomSample}(\mathcal{X}, K)$}
\STATE{\textbf{Given:} Initial evaluations $\mathcal{Y}_0 \leftarrow \{f(S, B) : (S, B) \in \mathcal{X}_0\}$}
\STATE{Initialize dataset $\mathcal{D}_0 \leftarrow \{(\mathcal{X}_0, \mathcal{Y}_0)\}$}
\FOR{$n = 1, ..., N$}
    \STATE{Update surrogate function $\hat{f}(S, B)$ using $\mathcal{D}_n$}
    \STATE{$(S_\text{next}, B_\text{next}) \leftarrow \arg\max \alpha(S, B | \mathcal{D}_n)$}
    \STATE{Train model $M \leftarrow \text{TrainModel}(S_\text{next}, B_\text{next})$}
    \STATE{Evaluate success rate $y_\text{next} \leftarrow \text{EvaluateModel}(M)$}
    \STATE{Update $\mathcal{D}_{n+1} \leftarrow \mathcal{D}_n \cup \{(S_\text{next}, B_\text{next}, y_\text{next})\}$}
    \IF{$\max \alpha(x; \mathcal{D}_n) < \epsilon$}
        \STATE{\textbf{break} (Stopping condition met: no promising regions remain to explore)}
    \ENDIF
    \IF{$y_\text{next} \geq \text{Success Threshold}$}
        \STATE{\textbf{return} Optimal parameters $(S_\text{next}, B_\text{next})$, Success rate $y_\text{next}$}
    \ENDIF
\ENDFOR
\STATE{\textbf{Output:} Best parameters $(S^*, B^*) = \arg\max \mathcal{Y}$, Best success rate $y^* = \max \mathcal{Y}$}
\end{algorithmic}
\end{algorithm}

% ================================================================%
\subsection{Validation Under Weather and Lighting Conditions}
% ================================================================%

After optimization, we evaluate the best object detection model under varying lighting conditions
(e.g., night) and weather conditions (e.g., heavy rain). We measure the landing success rate from 10 random initial vehicle positions that range from 0-40 m offset on the horizontal plane and 0-120 m in the vertical direction. For evaluating the confidence score of object detection we consider the score from the extreme offset position (40 m, 40 m and 120m in $\vect{x}$, $\vect{y}$ and $\vect{z}$ direction respectively). We imposed the same criteria for successful landing described in the simulation testing section. The landing success rate is then incorporated into Algorithm~\ref{alg:bayesian_optimization} to identify the best-performing model for these specific environmental states.
These optimizations generate hyperparameter spaces mapped to the landing success rate for each environmental condition.
We obtained models that consistently perform with more than 70\% landing success rates for each condition.
In addition to the best-performing model for each environmental condition, we identify a high-performing parameter space shared across different landing scenarios.
This common parameter space enables the training of a robust model that demonstrates improved performance
across all evaluated lighting and weather conditions. The final evaluation shows a 70\% landing success rate in clear day and night conditions and a 50\% landing success rate at night with rain conditions. An illustrative video of the final landing evaluation is available
here.\footnote{\url{https://www.youtube.com/watch?v=WAGicg_xGj8}}

%%%%%%%%%%%%%%%%%%%%%%%%%%%%%%%%%%%%%%%%%%%%%%%%%%%%%%
\section{Experiments and Result}
%%%%%%%%%%%%%%%%%%%%%%%%%%%%%%%%%%%%%%%%%%%%%%%%%%%%%%

% ===========================================
\subsection{Photorealistic simulator}
\begin{figure}[hbt!]
     \centering
     \begin{subfigure}[b]{0.49\columnwidth}
         \centering
         \includegraphics[width=\columnwidth]{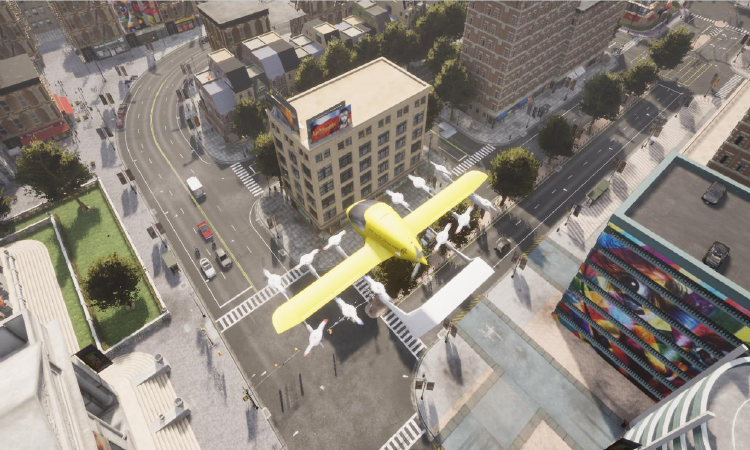}
         \caption{Photorealistic scene}
         \label{fig:vr}
     \end{subfigure}
     \hfill
     \begin{subfigure}[b]{0.49\columnwidth}
         \centering
         \includegraphics[width=\columnwidth]{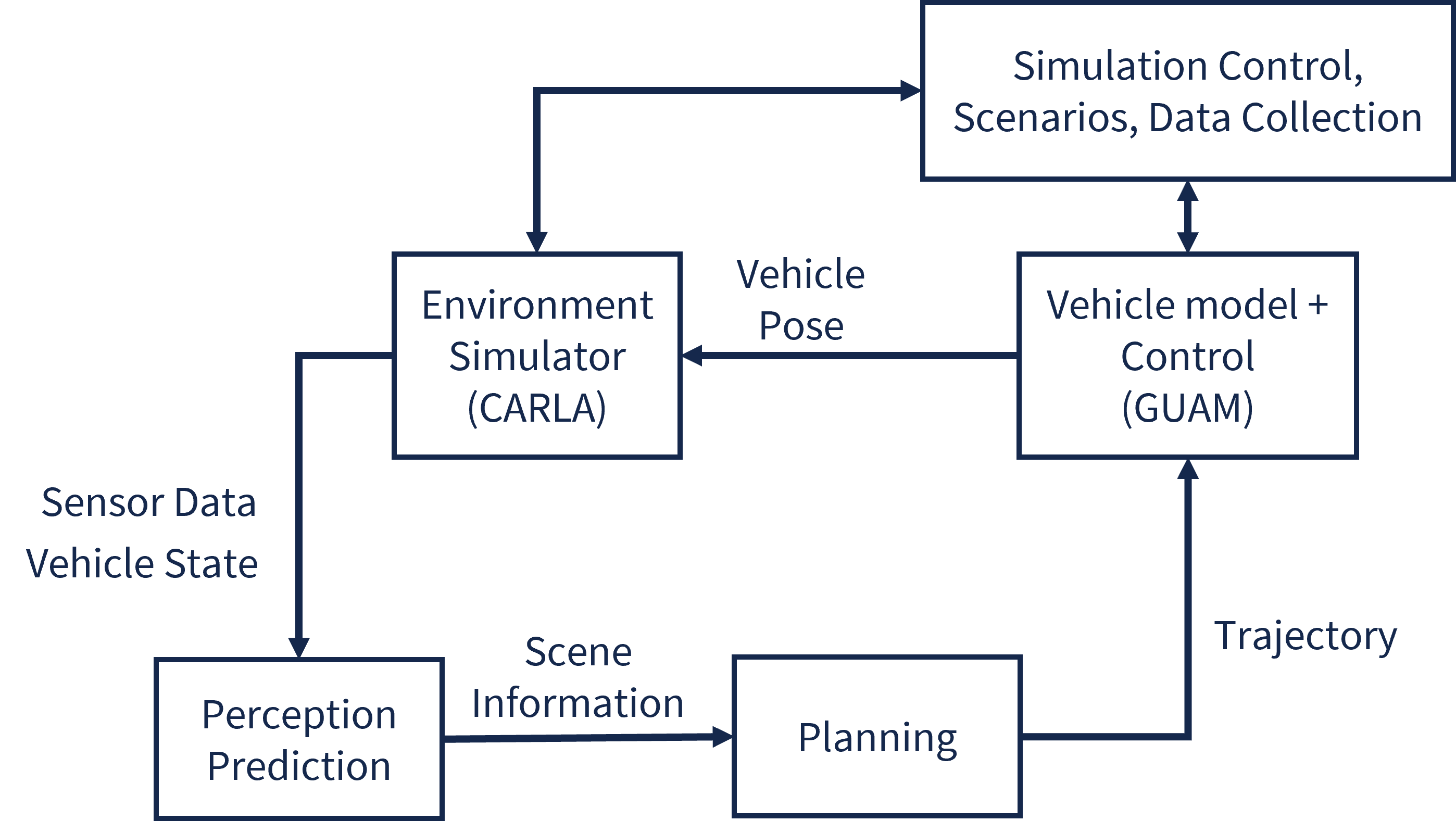}
         \caption{Modular structure of the simulator}
         \label{fig:carla_flow}
     \end{subfigure}
     \caption{Integration of perception, planning, and control with the simulator}
     \label{fig:overview}
\end{figure}

\begin{figure}[hbt!]
    \centering
    \begin{subfigure}[b]{.85\columnwidth}
        \centering
        \includegraphics[width=\columnwidth]{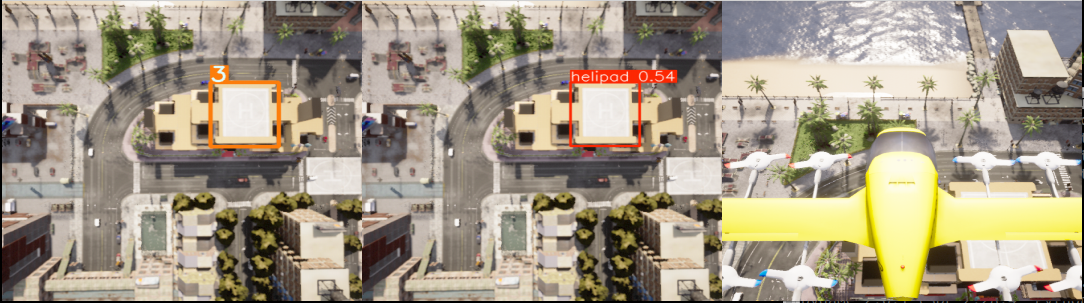}
        \caption{Clear Day}
        \label{fig:day_landing}
    \end{subfigure}
    \hfill
    \begin{subfigure}[b]{.85\columnwidth}
        \centering
        \includegraphics[width=\columnwidth]{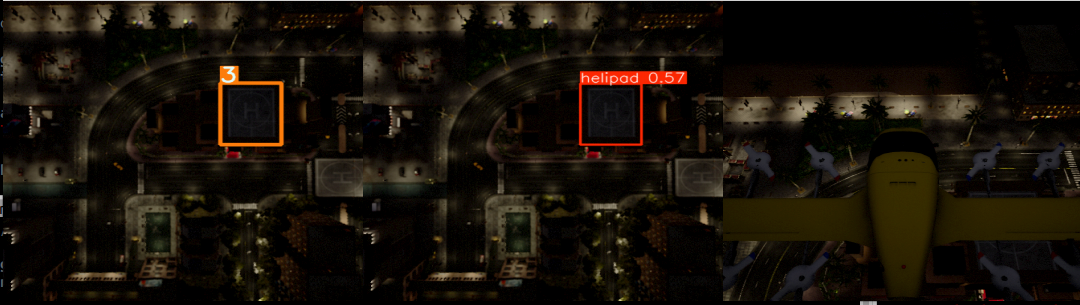}
        \caption{Night}
        \label{fig:night_landing}
    \end{subfigure}
    \hfill
    \begin{subfigure}[b]{.85\columnwidth}
        \centering
        \includegraphics[width=\columnwidth]{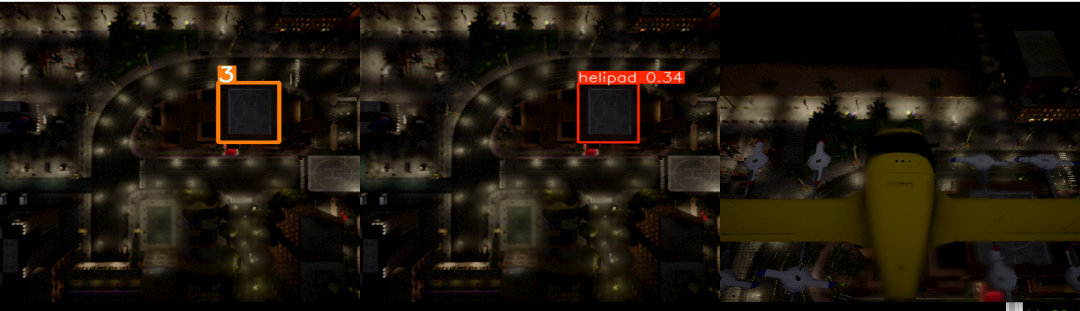}
        \caption{Night and Rain}
        \label{fig:night_rain_landing}
    \end{subfigure}
    \caption{Simulation for different landing scenarios.}

    \label{fig:variable_weather}
\end{figure}

% ===========================================
The photorealistic simulator in Figure~\ref{fig:vr} is developed using the existing autonomous car simulation application, \emph{CARLA}~\cite{Dosovitskiy17}. Using the \emph{Unreal Engine}~\cite{Unreal}, 3D rendering of an urban air mobility (\emph{UAM}) vehicle is added, as shown in Figure~\ref{fig:vr}. The simulator integrates vehicle dynamics, controller, planning, and perception. The data-exploration decision-making is implemented through Simulation Control, as shown in Figure~\ref{fig:carla_flow}. The simulation control chooses different weather conditions and lighting conditions, as shown in Figure~\ref{fig:variable_weather}.

% ===========================================
% ===========================================

   \begin{figure}[hbt!]
    \centering
    % First image
    \begin{subfigure}{0.45\columnwidth}
        \centering
        \includegraphics[width=\columnwidth]{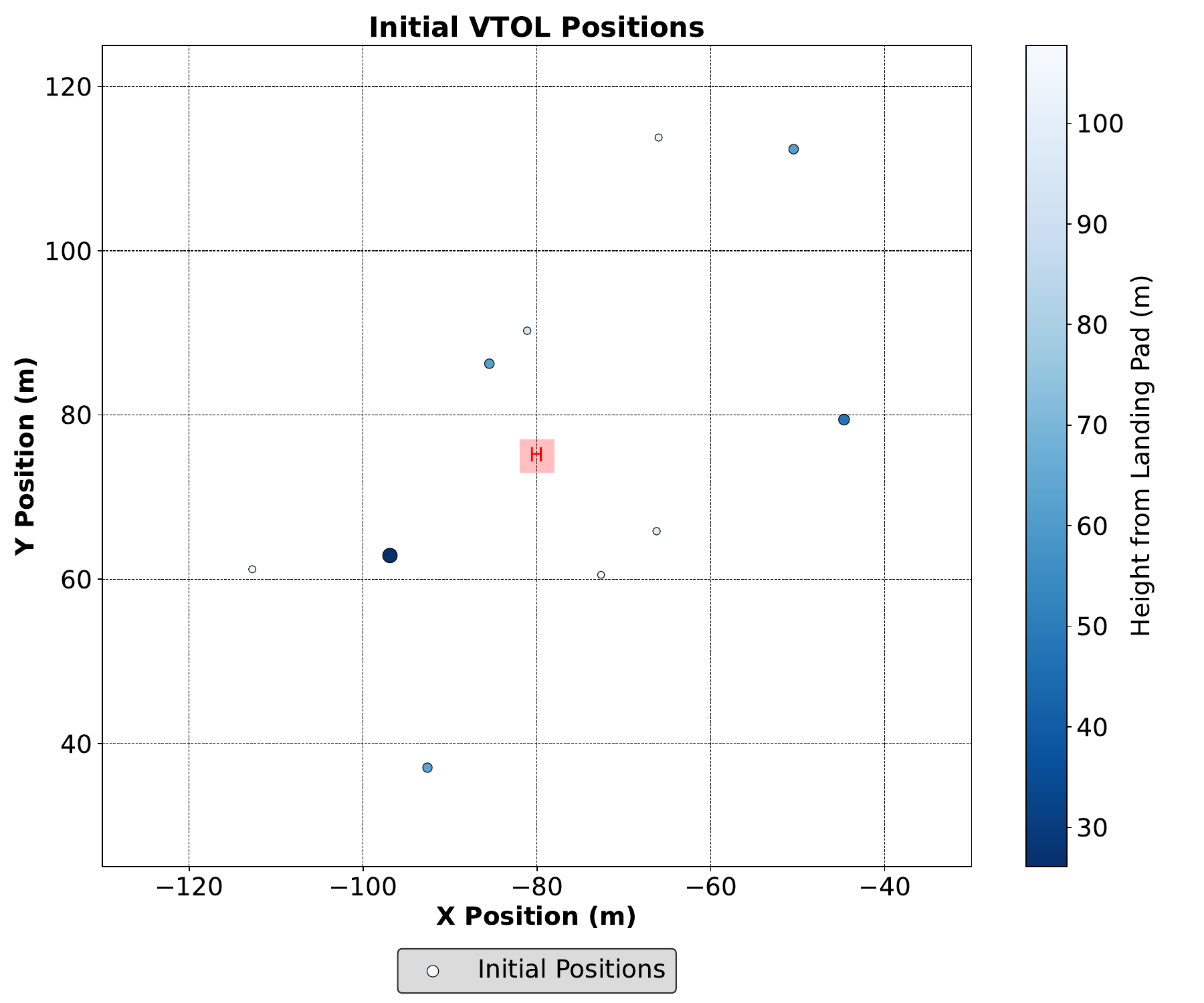}
        \caption{Initial Position of \emph{VTOL}
        }
        \label{fig:initial_pos_before}
    \end{subfigure}
    \hspace{0.02\textwidth} % Reduce space between images
    % Second image
    \begin{subfigure}{0.45\columnwidth}
        \centering
        \includegraphics[width=\columnwidth]{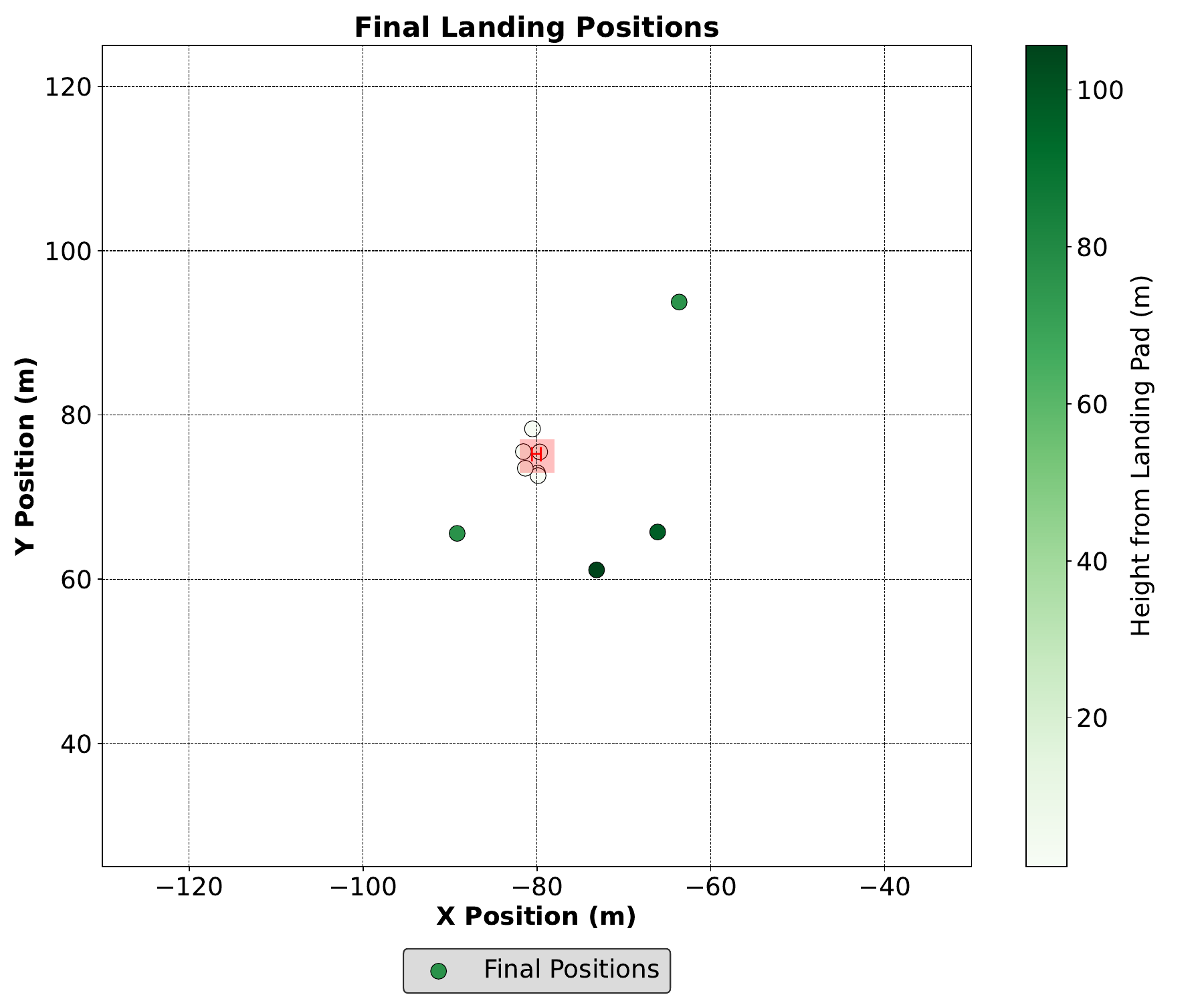}
        \caption{Final Landing Position}

        % \hyung{font size}

        \label{fig:final_pos_before}
    \end{subfigure}
    \caption{Landing Performance without Optimization}
    \label{fig:final_landing_before}
\end{figure}

   \begin{figure}[hbt!]
    \centering
    % First image
    \begin{subfigure}{0.45\columnwidth}
        \centering
        \includegraphics[width=\columnwidth]{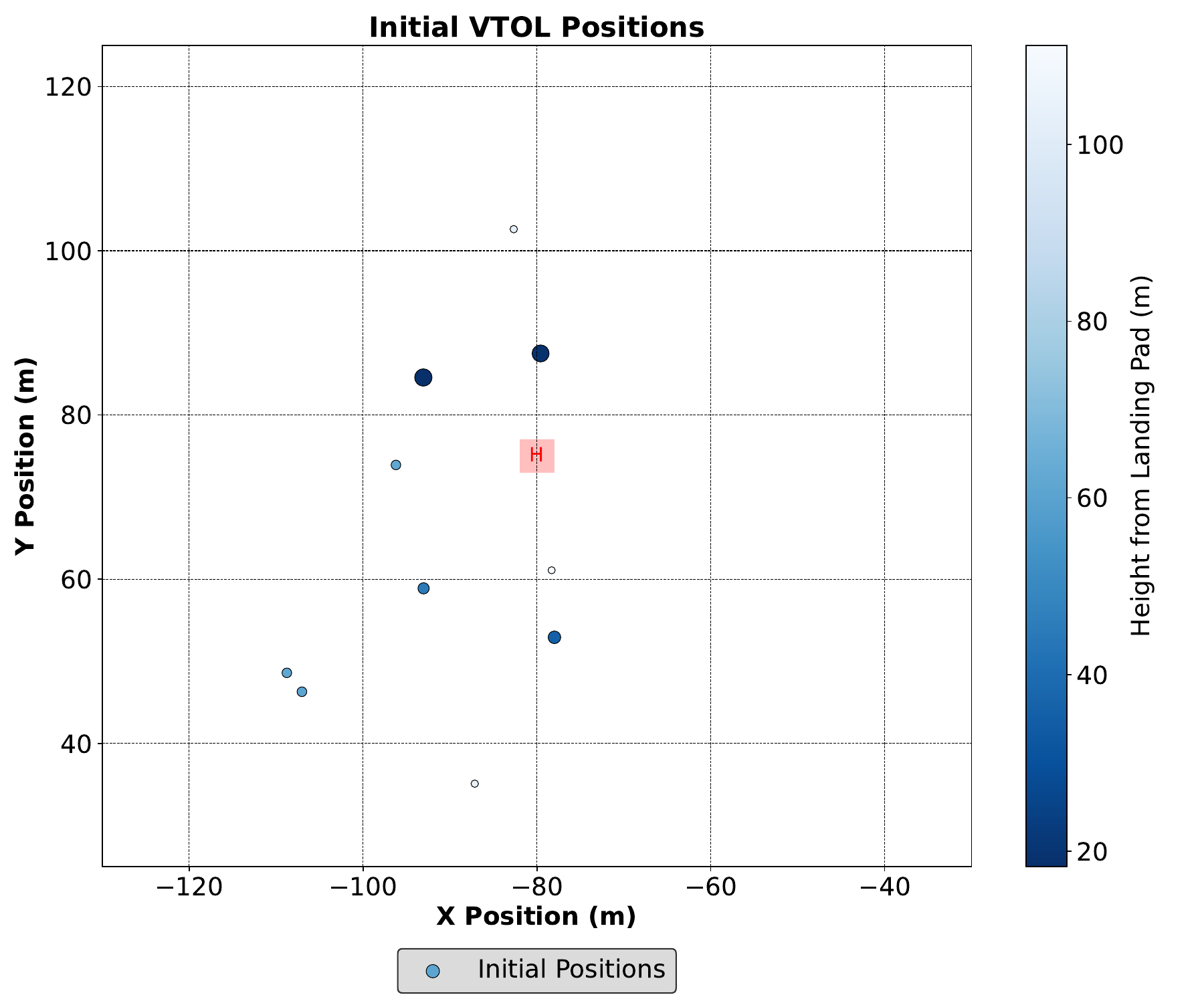}
        \caption{Initial Position of \emph{VTOL}}
        \label{fig:initail_pos_after}
    \end{subfigure}
    \hspace{0.02\textwidth} % Reduce space between images
    % Second image
    \begin{subfigure}{0.45\columnwidth}
        \centering
        \includegraphics[width=\columnwidth]{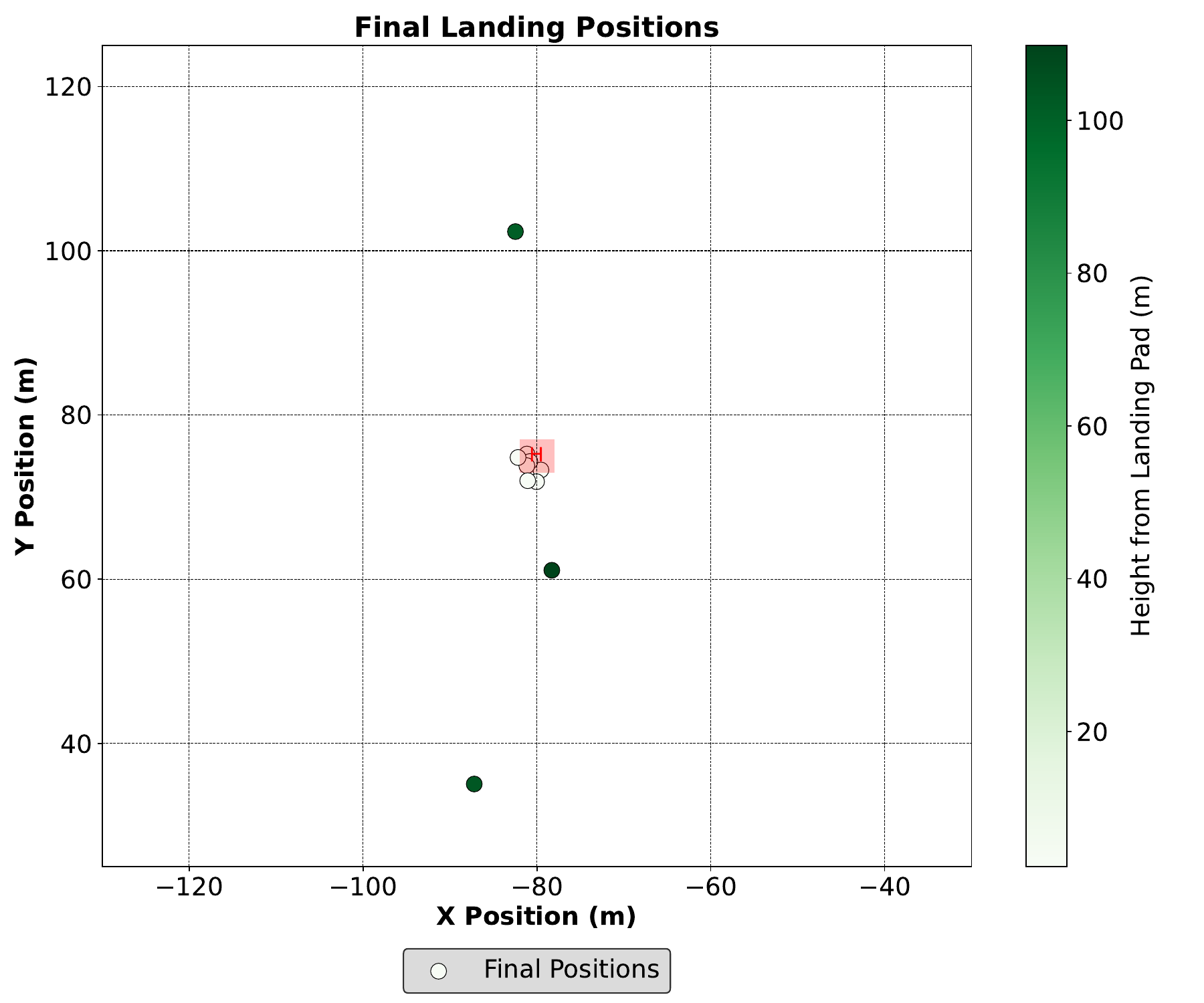}
        \caption{Final Landing Position}
        \label{final pos after}
    \end{subfigure}
    \caption{Landing Performance after Optimization (20\% Improvement in Landing Success Rate)}
        \label{fig:final_landing_after}
\end{figure}

\begin{figure}[hbt!]
    \centering
    \begin{subfigure}[b]{0.45\columnwidth}
        \centering
        \includegraphics[width=\columnwidth]{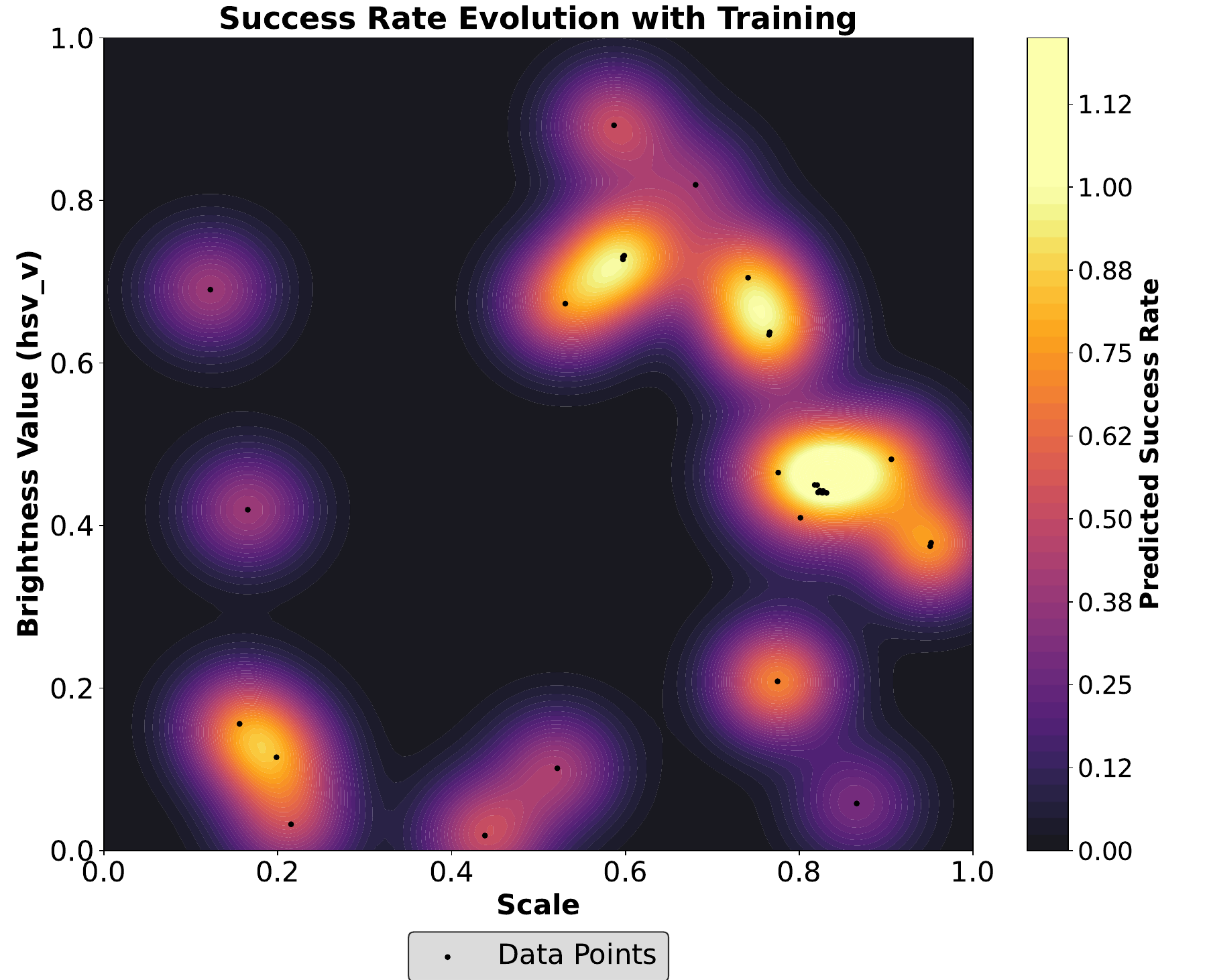}
        \caption{Clear Day}
         \label{fig:Optimization}
    \end{subfigure}
    \hfill
    \begin{subfigure}[b]{0.45\columnwidth}
        \centering
        \includegraphics[width=\columnwidth]{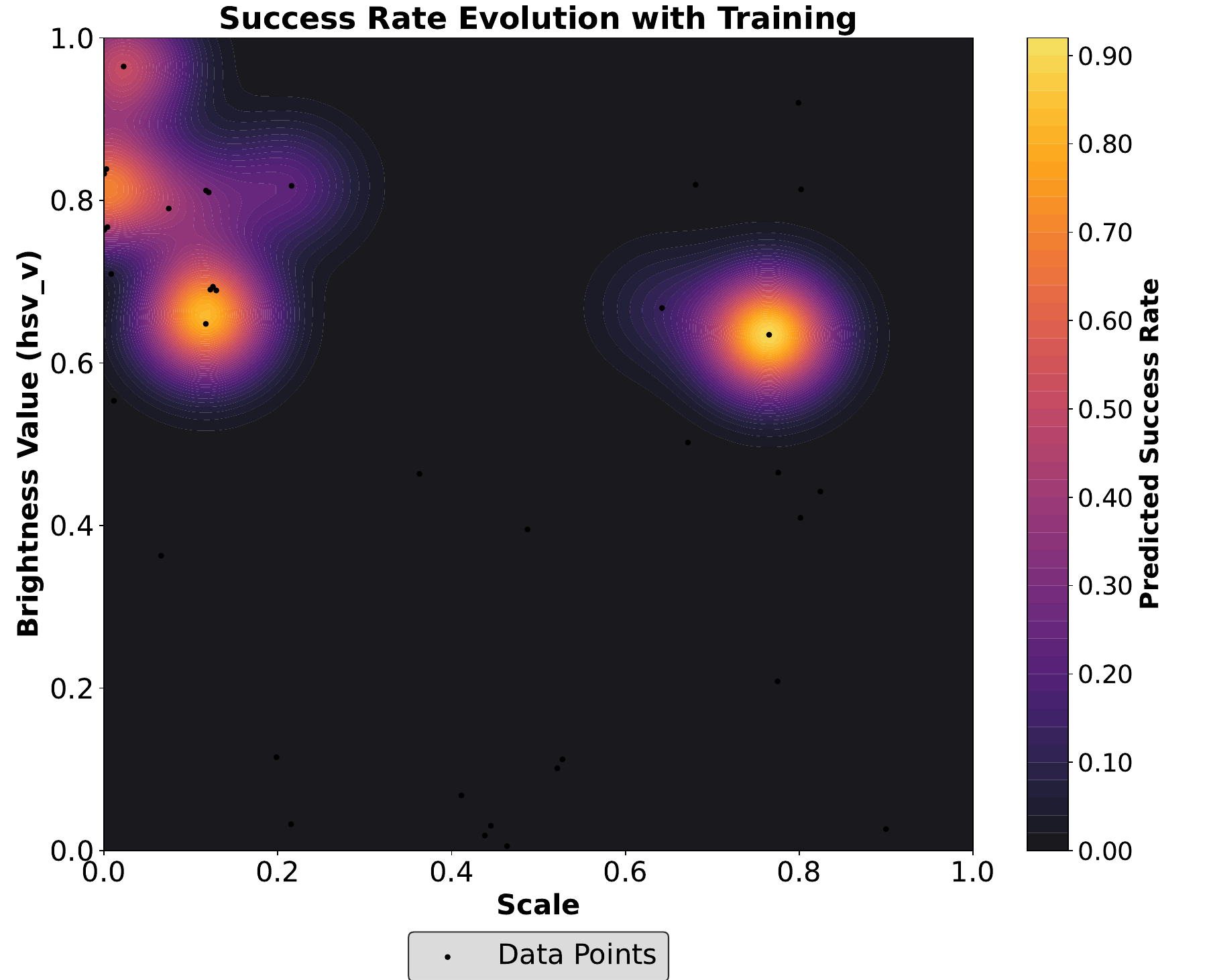}
        \caption{Clear Night}
         \label{fig:clear_night_bayes}
    \end{subfigure}
    \hfill
    \begin{subfigure}[b]{0.45\columnwidth}
        \centering
        \includegraphics[width=\columnwidth]{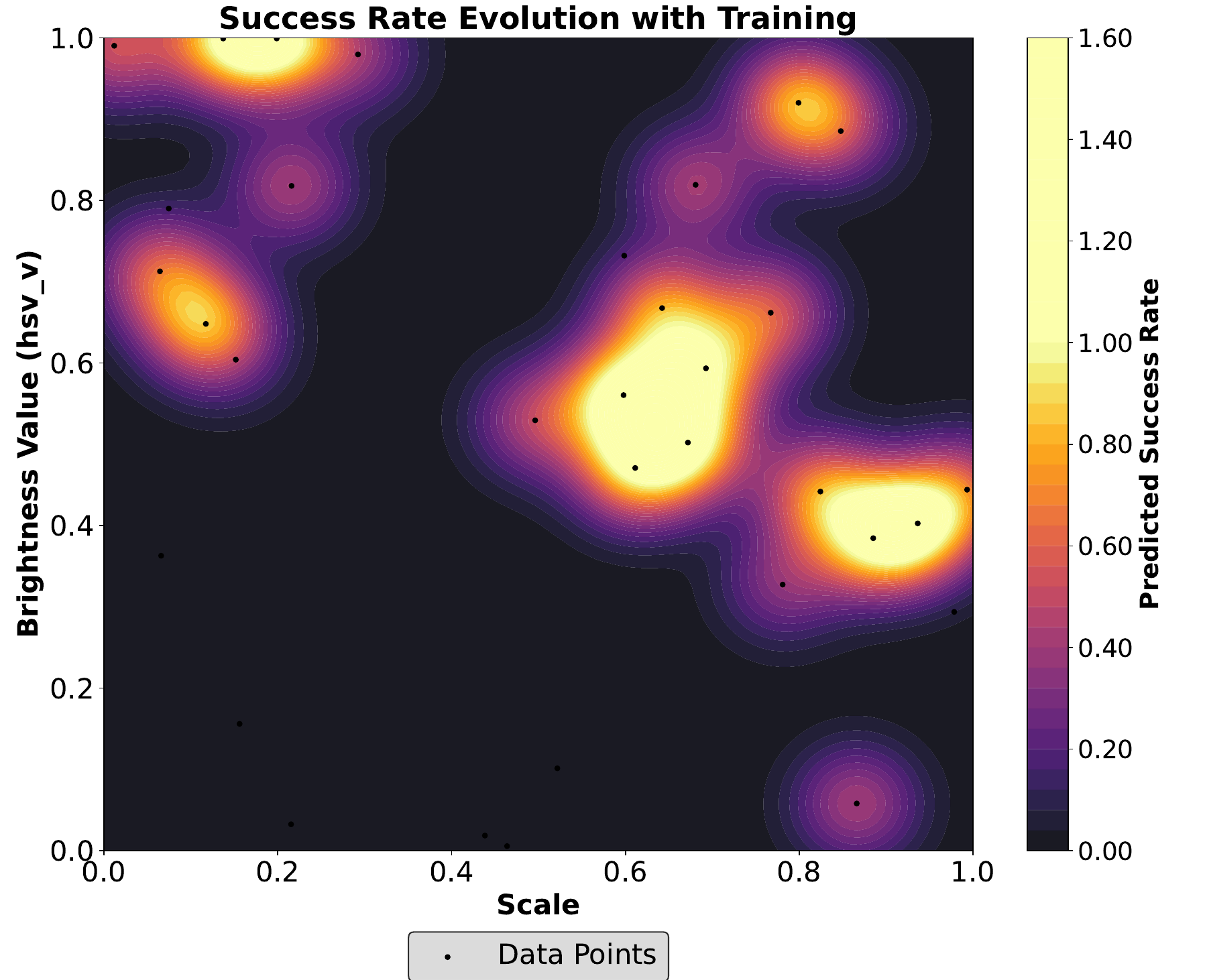}
        \caption{Night and Rain}
         \label{fig:night_rain_bayes}
    \end{subfigure}
    \caption{Best Performing Parameters for Landing}
    \label{Bayesian OPT}
\end{figure}

\subsection{Result}

The proposed framework was evaluated using key performance metrics, including mean average precision (mAP), confidence scores, and landing success rates, to assess its effectiveness in detecting helipads and guiding safe \emph{VTOL} landings.
The initial \emph{YOLOv8} model was trained using randomly selected augmentations (scale and brightness). During the evaluation, the model achieved an mAP (50-95) of 43\%, which is close to the benchmark mAP (50-95) for \emph{YOLOv8}~\cite{YOLOv8}, typically reported as 44.9\% for large-scale datasets, such as the \emph{COCO} dataset~\cite{lin2014microsoft}.
 The pre-trained model was integrated into \emph{CARLA} to assess the landing success rate from 10 randomly generated initial positions, as shown in Figure~\ref{fig:initial_pos_before}. The initial model achieved a success rate of 50\%, as illustrated in Figure~\ref{fig:final_pos_before}, under clear, bright weather conditions in \emph{CARLA}. This is a baseline evaluation of the model's performance.
The Bayesian optimization algorithm~\ref{alg:bayesian_optimization} uses the evaluation outcome, which consists of a pair of data augmentation hyperparameters (scale and brightness) and the corresponding success rate. The optimization process begins by collecting the evaluation outcomes, forming a closed loop of evaluation and optimization steps in each iteration.
After conducting 30 iterations, the landing success rate improved from the baseline of 50\% to 70\%, as shown in Figure~\ref{fig:final_landing_after}.
This experimental result demonstrates that the proposed Bayesian data augmentation approach can improve the landing success rate by identifying more suitable data augmentation hyperparameters.

To better understand the selection of the hyperparameters in the Bayesian optimization, we plotted a contour plot showing the estimated landing success rate for each pair of scale and brightness (estimated by the GP in Bayesian optimization), as shown in Figure~\ref{Bayesian OPT}. In Figure~\ref{Bayesian OPT}, we present the contours for different weather conditions: (a) clear day; (b) clear night; (c) night and rain.
As shown in Figure~\ref{fig:Optimization}, there are several pairs of scale and brightness augmentations that achieve the same landing success rate. Next, we selected one of these pairs and tested it under different lighting conditions (clear night). However, we achieved a zero landing success rate under this condition with the hyperparameters that worked well in the other conditions. This example demonstrates that Bayesian optimization, when performed under favorable conditions, may carry the risk of failure, in contrast to optimization conducted under more challenging or adverse conditions.
Next, we applied Bayesian optimization under two adverse conditions. First, we ran Algorithm~\ref{alg:bayesian_optimization} under clear night conditions, achieving a landing success rate of 70\% and identifying the corresponding best-performing parameters, as shown in Figure~\ref{fig:clear_night_bayes}. Then, we ran the algorithm under rain and night conditions to obtain the relevant set of hyperparameters, as shown in Figure~\ref{fig:night_rain_bayes}. As shown in Figure~\ref{Bayesian OPT}, the hyperparameters scale=0.77 and brightness=0.66 are jointly shared across all three environmental states as high-performing parameters. As a result, the model achieves a 70\% landing success rate in both day and night conditions, and a 50\% success rate in night with rain conditions. This suggests the potential of using adverse conditions to improve the robustness of the perception model. Furthermore, during the final validation of the perception model, we have observed the confidence scores for bounding boxes also showed enhanced reliability, even under varying lighting and weather conditions. The confidence score improved from 30\% to 50\% in clear day and from 0\% to 50\% in clear night landing scenario while evaluated from extreme offset position. These results emphasize the effectiveness of the proposed framework in improving \emph{VTOL}-specific perception and control tasks.

\section{Conclusion and Future Work}
This work introduces a novel framework for enhancing \emph{VTOL} perception systems that focuses on vision-guided landing performance. By integrating reproducible training of \emph{DNN} model, uncertainty estimation, simulation testing and Bayesian optimization, the framework addresses challenges related to dataset variability, environmental unpredictability and model refinement. The Bayesian data augmentation framework improved the landing success rate from the baseline of 50\% to 70\%, demonstrating its ability to adapt and enhance model performance under dynamic conditions. This framework highlights how domain-specific optimizations can bridge the gap between general-purpose object detection and specialized \emph{VTOL} tasks, outperforming typical \emph{YOLO} object detection benchmark models.

\begin{figure}[H]
    \centering
    \includegraphics[width=0.8\textwidth]{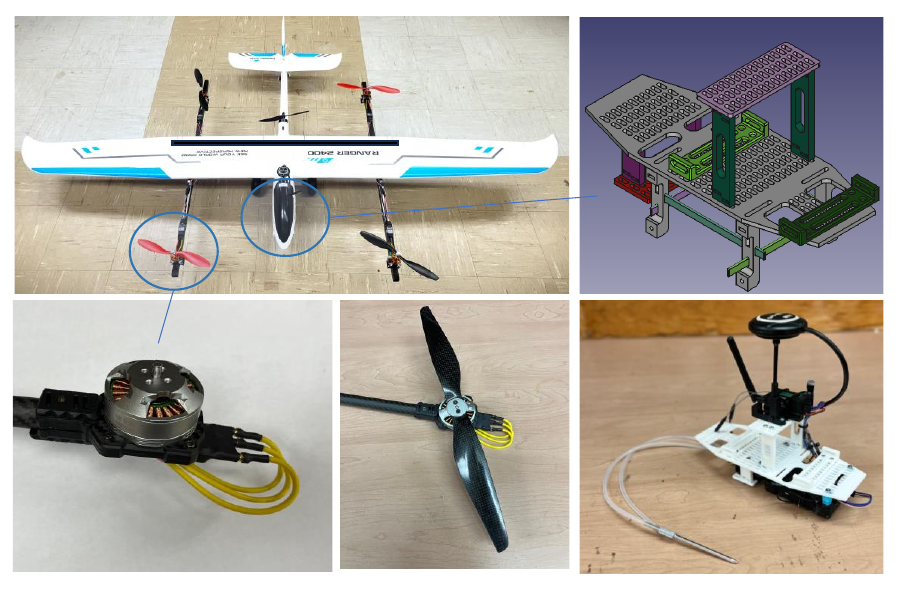}
    \caption{Prototype Design for Real Flight Experiment}
    \label{fig:prototype}
\end{figure}

Future work will focus on expanding the dataset, testing under additional environmental conditions, and transitioning to real-world \emph{VTOL} operations. By addressing these aspects, the framework has the potential to significantly advance the development of reliable, autonomous \emph{VTOL} systems capable of safe and efficient landings in complex environments. Building on these promising simulation results, our next step is to test the model in real-flight scenarios using the platform currently under development, as shown in Figure~\ref{fig:prototype}.

\section*{Acknowledgments}

This material is based upon work supported by
the National Aeronautics and Space Administration (NASA) under the cooperative agreement 80NSSC20M0229 and University Leadership Initiative grant no. 80NSSC22M0070, and
the National Science Foundation (NSF) under grant no. CNS 1932529 and ECCS 2311085.

Any opinions, findings, conclusions or recommendations expressed in this material
are those of the authors and do not necessarily reflect
the views of the sponsors.%

\bibliography{mybib}

\begin{thebibliography}{45}
\newcommand{\enquote}[1]{``#1''}
\providecommand{\natexlab}[1]{#1}
\providecommand{\url}[1]{\texttt{#1}}
\providecommand{\urlprefix}{URL }
\expandafter\ifx\csname urlstyle\endcsname\relax
  \providecommand{\doi}[1]{\discretionary{}{}{}https://doi.org/#1}\else
  \providecommand{\doi}[1]{\discretionary{}{}{}\urlstyle{rm}\url{https://doi.org/#1}}\fi

\bibitem[{Grigorescu et~al.(2020)Grigorescu, Trasnea, Cocias, and Macesanu}]{grigorescu2020survey}
Grigorescu, S., Trasnea, B., Cocias, T., and Macesanu, G., \enquote{A survey of deep learning techniques for autonomous driving,} \emph{Journal of Field Robotics}, Vol.~37, No.~3, 2020, pp. 362--386.

\bibitem[{Redmon et~al.(2016)Redmon, Divvala, Girshick, and Farhadi}]{redmon2016}
Redmon, J., Divvala, S., Girshick, R., and Farhadi, A., \enquote{You only look once: Unified, real-time object detection,} \emph{Proceedings of the IEEE conference on computer vision and pattern recognition}, 2016, pp. 779--788.

\bibitem[{Redmon and Farhadi(2017)}]{redmon2017yolo9000}
Redmon, J., and Farhadi, A., \enquote{YOLO9000: better, faster, stronger,} \emph{Proceedings of the IEEE conference on computer vision and pattern recognition}, 2017, pp. 7263--7271.

\bibitem[{Ultralytics(2020)}]{ultralytics2020yolov5}
Ultralytics, \enquote{YOLOv5: You Only Look Once version 5,} \url{https://github.com/ultralytics/yolov5}, 2020.

\bibitem[{Ultralytics(2023{\natexlab{a}})}]{ultralytics2023yolov8}
Ultralytics, \enquote{YOLOv8: You Only Look Once version 8,} \url{https://github.com/ultralytics/ultralytics}, 2023{\natexlab{a}}.

\bibitem[{Gupta et~al.(2024)Gupta, Kotlyar, Andreasson, and Lilienthal}]{Gupta_2024_WACV}
Gupta, H., Kotlyar, O., Andreasson, H., and Lilienthal, A.~J., \enquote{Robust Object Detection in Challenging Weather Conditions,} \emph{Proceedings of the IEEE/CVF Winter Conference on Applications of Computer Vision (WACV)}, 2024, pp. 7523--7532.

\bibitem[{Docca and Torculas(2023)}]{nvidia_synthetic}
Docca, A., and Torculas, M., \enquote{How to Train a Defect Detection Model Using Synthetic Data with NVIDIA Omniverse Replicator,} \url{https://developer.nvidia.com/blog/how-to-train-a-defect-detection-model-using-synthetic-data-with-nvidia-omniverse-replicator}, 2023.
\newblock Accessed: 2024-05-22.

\bibitem[{Dallel et~al.(2023)Dallel, Havard, Dupuis, and Baudry}]{dallel2023digital}
Dallel, M., Havard, V., Dupuis, Y., and Baudry, D., \enquote{Digital twin of an industrial workstation: A novel method of an auto-labeled data generator using virtual reality for human action recognition in the context of human--robot collaboration,} \emph{Engineering applications of artificial intelligence}, Vol. 118, 2023, p. 105655.

\bibitem[{Technologies(2023)}]{unity_autonomous_vehicle_training}
Technologies, U., \enquote{Efficient Development of Simulated Environments for Autonomous Vehicle Training,} \url{https://unity.com/how-to/simulated-environments-for-autonomous-vehicle-training}, 2023.
\newblock Accessed: 2023-08-22.

\bibitem[{Thompson et~al.(2020)Thompson, Greenewald, Lee, and Manso}]{thompson2020computational}
Thompson, N.~C., Greenewald, K., Lee, K., and Manso, G.~F., \enquote{The computational limits of deep learning,} \emph{arXiv preprint arXiv:2007.05558}, Vol.~10, 2020.

\bibitem[{Frazier(2018)}]{frazier2018tutorial}
Frazier, P.~I., \enquote{A tutorial on Bayesian optimization,} \emph{arXiv preprint arXiv:1807.02811}, 2018.

\bibitem[{Victoria and Maragatham(2021)}]{victoria2021automatic}
Victoria, A.~H., and Maragatham, G., \enquote{Automatic Tuning of Hyperparameters Using Bayesian Optimization,} \emph{Evolving Systems}, Vol.~12, 2021, pp. 217--223.

\bibitem[{Dosovitskiy et~al.(2017)Dosovitskiy, Ros, Codevilla, Lopez, and Koltun}]{Dosovitskiy17}
Dosovitskiy, A., Ros, G., Codevilla, F., Lopez, A., and Koltun, V., \enquote{{CARLA}: {An} Open Urban Driving Simulator,} \emph{Proceedings of the 1st Annual Conference on Robot Learning}, 2017, pp. 1--16.

\bibitem[{NASA(2024)}]{GUAM2024}
NASA, \enquote{Generic Urban Air Mobility (GUAM),} \url{https://github.com/nasa/Generic-Urban-Air-Mobility-GUAM}, 2024.
\newblock Accessed: November 23, 2024.

\bibitem[{Xie et~al.(2019)Xie, Tan, Gong, Wang, Yuille, and Le}]{xie2019ada}
Xie, C., Tan, M., Gong, B., Wang, J., Yuille, A., and Le, Q.~V., \enquote{ADA: Adversarial Data Augmentation for Object Detection,} \emph{Proceedings of the IEEE/CVF International Conference on Computer Vision (ICCV)}, IEEE, 2019, pp. 7364--7373.

\bibitem[{Mishra et~al.(2024)Mishra, Sieb, Abbeel, and Chen}]{mishra2024closing}
Mishra, N., Sieb, M., Abbeel, P., and Chen, X., \enquote{Closing the Visual Sim-to-Real Gap with Object-Composable NeRFs,} \emph{arXiv preprint arXiv:2403.04114}, 2024.
\newblock \urlprefix\url{https://arxiv.org/abs/2403.04114}.

\bibitem[{Paul et~al.(2021)Paul, Ganguli, and Dziugaite}]{paul2021deep}
Paul, M., Ganguli, S., and Dziugaite, G.~K., \enquote{Deep learning on a data diet: Finding important examples early in training,} \emph{Advances in Neural Information Processing Systems}, Vol.~34, 2021, pp. 20596--20607.

\bibitem[{Mindermann et~al.(2022)Mindermann, Brauner, Razzak, Sharma, Kirsch, Xu, H{\"o}ltgen, Gomez, Morisot, Farquhar et~al.}]{mindermann2022prioritized}
Mindermann, S., Brauner, J.~M., Razzak, M.~T., Sharma, M., Kirsch, A., Xu, W., H{\"o}ltgen, B., Gomez, A.~N., Morisot, A., Farquhar, S., et~al., \enquote{Prioritized training on points that are learnable, worth learning, and not yet learnt,} \emph{International Conference on Machine Learning}, PMLR, 2022, pp. 15630--15649.

\bibitem[{Guo et~al.(2022)Guo, Zhao, and Bai}]{guo2022deepcore}
Guo, C., Zhao, B., and Bai, Y., \enquote{Deepcore: A comprehensive library for coreset selection in deep learning,} \emph{International Conference on Database and Expert Systems Applications}, Springer, 2022, pp. 181--195.

\bibitem[{Sorscher et~al.(2022)Sorscher, Geirhos, Shekhar, Ganguli, and Morcos}]{sorscher2022beyond}
Sorscher, B., Geirhos, R., Shekhar, S., Ganguli, S., and Morcos, A., \enquote{Beyond neural scaling laws: beating power law scaling via data pruning,} \emph{Advances in Neural Information Processing Systems}, Vol.~35, 2022, pp. 19523--19536.

\bibitem[{Deng et~al.(2024)Deng, Cui, and Zhu}]{deng2024towards}
Deng, Z., Cui, P., and Zhu, J., \enquote{Towards Accelerated Model Training via Bayesian Data Selection,} \emph{Advances in Neural Information Processing Systems}, Vol.~36, 2024.

\bibitem[{Shahriari et~al.(2015)Shahriari, Swersky, Wang, Adams, and De~Freitas}]{shahriari2015taking}
Shahriari, B., Swersky, K., Wang, Z., Adams, R.~P., and De~Freitas, N., \enquote{Taking the human out of the loop: A review of Bayesian optimization,} \emph{Proceedings of the IEEE}, Vol. 104, No.~1, 2015, pp. 148--175.

\bibitem[{Snoek et~al.(2012)Snoek, Larochelle, and Adams}]{snoek2012practical}
Snoek, J., Larochelle, H., and Adams, R.~P., \enquote{Practical bayesian optimization of machine learning algorithms,} \emph{Advances in neural information processing systems}, Vol.~25, 2012.

\bibitem[{Fern{\'a}ndez~Casta{\~n}o et~al.(2024)Fern{\'a}ndez~Casta{\~n}o, Patton, Yoon, and Voulgaris}]{fernandez2024uncertainty}
Fern{\'a}ndez~Casta{\~n}o, A., Patton, C., Yoon, H.~J., and Voulgaris, P., \enquote{Uncertainty Quantification-Based Switching Control Method for Vision-Based Object Tracking in Unmanned Aerial Vehicles,} \emph{AIAA SCITECH 2024 Forum}, 2024, p. 0516.

\bibitem[{Cao and Hovakimyan(2008)}]{cao2008design}
Cao, C., and Hovakimyan, N., \enquote{Design and analysis of a novel $\mathcal{L}_1$ adaptive control architecture with guaranteed transient performance,} \emph{IEEE Transactions on Automatic Control}, Vol.~53, No.~2, 2008, pp. 586--591.

\bibitem[{Hose et~al.(2024)Hose, Brunzema, von Rohr, Gr{\"a}fe, Schoellig, and Trimpe}]{hose2024finetuning}
Hose, H., Brunzema, P., von Rohr, A., Gr{\"a}fe, A., Schoellig, A.~P., and Trimpe, S., \enquote{Fine-Tuning of Neural Network Approximate {MPC} without Retraining via Bayesian Optimization,} \emph{CoRL Workshop on Safe and Robust Robot Learning for Operation in the Real World}, 2024.
\newblock \urlprefix\url{https://openreview.net/forum?id=lSah6an1Ar}.

\bibitem[{Ostafew et~al.(2014)Ostafew, Schoellig, and Barfoot}]{6907444}
Ostafew, C.~J., Schoellig, A.~P., and Barfoot, T.~D., \enquote{Learning-based nonlinear model predictive control to improve vision-based mobile robot path-tracking in challenging outdoor environments,} \emph{2014 IEEE International Conference on Robotics and Automation (ICRA)}, 2014, pp. 4029--4036.
\newblock \doi{10.1109/ICRA.2014.6907444}.

\bibitem[{Terven and Cordova-Esparza(2023)}]{terven2023yolo}
Terven, J., and Cordova-Esparza, D., \enquote{A Comprehensive Review of YOLO Architectures in Computer Vision: From YOLOv1 to YOLOv8 and YOLO-NAS,} \emph{arXiv preprint arXiv:2304.00501}, 2023.
\newblock \urlprefix\url{https://arxiv.org/abs/2304.00501}.

\bibitem[{Bitoun and Winkler(2020)}]{bitoun2020helipadcat}
Bitoun, J., and Winkler, S., \enquote{HelipadCat: Categorised Helipad Image Dataset and Detection Method,} \emph{Proceedings of the IEEE Region 10 Conference (TENCON)}, IEEE, 2020, pp. 685--689.

\bibitem[{Ultralytics(2024)}]{ultralytics2024}
Ultralytics, \emph{Ultralytics YOLO Documentation: Configuration}, 2024.
\newblock \urlprefix\url{https://docs.ultralytics.com/usage/cfg/}, accessed: 2024-12-01.

\bibitem[{Tatman et~al.(2018{\natexlab{a}})Tatman, Vanderplas, and Dane}]{tatman2018reproducibility}
Tatman, R., Vanderplas, J., and Dane, K., \enquote{State of the Art: Reproducibility in Artificial Intelligence,} \emph{Proceedings of the 2018 Conference on Empirical Methods in Natural Language Processing}, 2018{\natexlab{a}}, pp. 1--10.

\bibitem[{Tatman et~al.(2018{\natexlab{b}})Tatman, VanderPlas, and Dane}]{tatman2018practical}
Tatman, R., VanderPlas, J., and Dane, S., \enquote{A practical taxonomy of reproducibility for machine learning research,} , 2018{\natexlab{b}}.

\bibitem[{Ultralytics(2023{\natexlab{b}})}]{YOLOv8}
Ultralytics, \enquote{YOLOv8,} \url{https://github.com/ultralytics/ultralytics}, 2023{\natexlab{b}}.

\bibitem[{Lin et~al.(2014)Lin, Maire, Belongie, Hays, Perona, Ramanan, Doll{\'a}r, and Zitnick}]{lin2014microsoft}
Lin, T.-Y., Maire, M., Belongie, S., Hays, J., Perona, P., Ramanan, D., Doll{\'a}r, P., and Zitnick, C.~L., \enquote{Microsoft coco: Common objects in context,} \emph{European conference on computer vision}, Springer, 2014, pp. 740--755.

\bibitem[{Breiman(2001)}]{breiman2001random}
Breiman, L., \enquote{Random forests,} \emph{Machine Learning}, Vol.~45, No.~1, 2001, pp. 5--32.

\bibitem[{H{\"u}llermeier and Waegeman(2019)}]{huellermeier2019aleatoric}
H{\"u}llermeier, E., and Waegeman, W., \enquote{Aleatoric and Epistemic Uncertainty in Machine Learning: An Introduction to Concepts and Methods,} \emph{arXiv preprint arXiv:1910.09457}, 2019.
\newblock \urlprefix\url{https://arxiv.org/abs/1910.09457}.

\bibitem[{Unr(2021)}]{Unreal}
\enquote{{Unreal Engine},} , 2021.
\newblock \urlprefix\url{https://www.unrealengine.com/}, accessed: 2021-12-01.

\bibitem[{So(2023)}]{jaxguam2023}
So, O., \enquote{JAX-GUAM: General Uncertainty Approximation Models,} \url{https://github.com/oswinso/jax_guam}, 2023.

\bibitem[{Paszke et~al.(2019)Paszke, Gross, Massa, Lerer, Bradbury, Chanan, Killeen, Lin, Gimelshein, Antiga, Desmaison, Kopf, Yang, DeVito, Raison, Tejani, Chilamkurthy, Steiner, Fang, Bai, and Chintala}]{NEURIPS2019_9015}
Paszke, A., Gross, S., Massa, F., Lerer, A., Bradbury, J., Chanan, G., Killeen, T., Lin, Z., Gimelshein, N., Antiga, L., Desmaison, A., Kopf, A., Yang, E., DeVito, Z., Raison, M., Tejani, A., Chilamkurthy, S., Steiner, B., Fang, L., Bai, J., and Chintala, S., \enquote{PyTorch: An Imperative Style, High-Performance Deep Learning Library,} \emph{Advances in Neural Information Processing Systems 32}, Curran Associates, Inc., 2019, pp. 8024--8035.
\newblock \urlprefix\url{http://papers.neurips.cc/paper/9015-pytorch-an-imperative-style-high-performance-deep-learning-library.pdf}.

\bibitem[{{Stanford Artificial Intelligence Laboratory et al.}(2018)}]{ros}
{Stanford Artificial Intelligence Laboratory et al.}, \enquote{Robotic Operating System,} , 2018.
\newblock \urlprefix\url{https://www.ros.org}.

\bibitem[{Bewley et~al.(2016)Bewley, Ge, Ott, Ramos, and Upcroft}]{bewley2016sort}
Bewley, A., Ge, Z., Ott, L., Ramos, F., and Upcroft, B., \enquote{Simple online and realtime tracking,} \emph{2016 IEEE international conference on image processing (ICIP)}, IEEE, 2016, pp. 3464--3468.

\bibitem[{{\AA}str{\"o}m and Murray(2021)}]{aastrom2021feedback}
{\AA}str{\"o}m, K.~J., and Murray, R., \emph{Feedback systems: an introduction for scientists and engineers}, Princeton university press, 2021.

\bibitem[{{\c{S}}umnu et~al.(2017){\c{S}}umnu, G{\"u}zelbey, and {\c{C}}akir}]{csumnu2017simulation}
{\c{S}}umnu, A., G{\"u}zelbey, {\.I}.~H., and {\c{C}}akir, M.~V., \enquote{Simulation and PID control of a Stewart platform with linear motor,} \emph{Journal of mechanical science and technology}, Vol.~31, 2017, pp. 345--356.

\bibitem[{Rasmussen and Nickisch(2010)}]{rasmussen2010gaussian}
Rasmussen, C.~E., and Nickisch, H., \enquote{Gaussian processes for machine learning (GPML) toolbox,} \emph{The Journal of Machine Learning Research}, Vol.~11, 2010, pp. 3011--3015.

\bibitem[{Nogueira(2014--)}]{BayesinaOptPython}
Nogueira, F., \enquote{{Bayesian Optimization}: Open source constrained global optimization tool for {Python},} , 2014--.
\newblock \urlprefix\url{https://github.com/bayesian-optimization/BayesianOptimization}.

\end{thebibliography}

\end{document}